\documentclass[twoside,11pt]{article}
\usepackage{jair, theapa, rawfonts}
\usepackage{amsfonts} 
\usepackage{amssymb}
\usepackage{booktabs}
\usepackage{graphicx}
\usepackage{float}
\usepackage[table]{xcolor}
\usepackage{amsmath}
\usepackage{mathtools}
\usepackage{enumitem}
\usepackage{url}
\usepackage{xr}
\usepackage{microtype}
\usepackage{multirow}
\usepackage[ruled,vlined,noend,linesnumbered]{algorithm2e}
\jairheading{1}{2022}{1-32}{--/--}{--/--}
\ShortHeadings{FlexiBERT: Are Current Transformer Architectures too Homogeneous and Rigid?}
{Tuli, Dedhia, Tuli, \& Jha}
\firstpageno{1}

\newcommand{\perf}{0.4\% }
\newcommand{\parambertmini}{3\% }
\newcommand{\perfbertmini}{8.9\% }
\newcommand{\perfhomo}{3\% }
\newcommand{\paramhomo}{2.6$\times$ }
\newcommand{\nasperf}{13\%}
\newcommand{\largeperf}{5.7\% }

\begin{document}

\title{FlexiBERT: Are Current Transformer Architectures too Homogeneous and Rigid?}

\author{\name Shikhar Tuli \email stuli@princeton.edu \\
       \name Bhishma Dedhia \email bdedhia@princeton.edu \\
       \addr Dept. of Electrical \& Computer Engineering, Princeton University \\
       Princeton, NJ 08544 USA
       \AND
       \name Shreshth Tuli \email s.tuli20@imperial.ac.uk \\
       \addr Department of Computing, Imperial College London \\
       London, SW7 2AZ UK
       \AND
       \name Niraj K. Jha \email jha@princeton.edu \\
       \addr Dept. of Electrical \& Computer Engineering, Princeton University \\
       Princeton, NJ 08544 USA}

% For research notes, remove the comment character in the line below.
% \researchnote

\maketitle

\begin{abstract}
% Setting the context
The existence of a plethora of language models makes the problem of selecting the 
best one for a custom task challenging.
% Hinting the challenge
Most state-of-the-art methods leverage transformer-based models (e.g., BERT) 
or their variants. Training such models and exploring their hyperparameter space,
however, is computationally expensive. 
% Setting the domain
Prior work proposes several neural architecture search (NAS) methods that employ performance
predictors (e.g., surrogate models) to address this
issue; however, analysis has been limited to homogeneous models that use fixed dimensionality 
throughout the network. This leads to sub-optimal architectures.
% Contribution
To address this limitation, we propose a suite of \emph{heterogeneous} and \emph{flexible} models, namely FlexiBERT, that 
have varied encoder layers with a diverse set of possible operations and different hidden dimensions. For better-posed 
surrogate modeling in this expanded design space, we propose a new graph-similarity-based 
embedding scheme. We also propose a novel NAS policy, called BOSHNAS, that leverages this 
new scheme, Bayesian modeling, and second-order optimization, to quickly train and use a 
neural surrogate model to converge to the optimal architecture.
% Evidence
A comprehensive set of experiments shows that the proposed policy, when applied to the \mbox{FlexiBERT} 
design space, pushes the performance frontier upwards compared to traditional models. FlexiBERT-Mini, one of our proposed models, has \parambertmini fewer parameters than BERT-Mini 
and achieves \perfbertmini higher GLUE score. A FlexiBERT model with equivalent performance as the best homogeneous model achieves \paramhomo smaller size. FlexiBERT-Large, another proposed model, achieves state-of-the-art results, outperforming the baseline models by at least \largeperf on the GLUE benchmark.
\end{abstract}

\section{Introduction}

In recent years, self-attention (SA)-based transformer models~\shortcite{vaswani,bert} have achieved
state-of-the-art results on tasks that span the natural language processing (NLP) domain. This burgeoning success has largely been driven by large-scale pre-training datasets, increasing computational power, and robust training techniques~\shortcite{roberta}. A challenge that remains is \emph{efficient} optimal model selection for a specific task and a set of user requirements. In this context, only those models should be trained that have the maximum \emph{predicted} performance.  This falls in the domain of neural architecture search (NAS)~\shortcite{nas_google}.

\subsection{Challenges}

The design space of transformer models is vast. Several models have been 
proposed in the past after rigorous search. Popular models include BERT, XLM,
XLNet, BART, ConvBERT, and FNet~\shortcite{bert,xlm,xlnet,bart,convbert,fnet}. Transformer design involves a choice 
of several hyperparameters, including the number of layers, size of 
hidden embeddings, number of attention heads, and size of the hidden layer in the 
feed-forward network~\shortcite{schubert}. This leads to an exponential increase in the design space, making a brute-force approach to explore the design space computationally infeasible~\shortcite{nasbench}. The aim is to converge to an optimal
model as quickly as possible, by testing the lowest possible 
number of datapoints~\shortcite{pham2018efficient}. Moreover, model performance may not be deterministic, requiring heteroscedastic modeling \shortcite{ru2020neural}. 

\subsection{Existing solutions}

Recent NAS advancements use various techniques to explore 
and optimize different models in the deep learning domain, from image recognition to speech 
recognition and machine translation \shortcite{nas_google,interspeech}. In the computer-vision 
domain, many convolutional neural network (CNN) architectures have been designed using 
various search approaches, such as genetic algorithms, reinforcement learning, structure adaptation,
etc. Some even introduce new basic operations \shortcite{shufflenet} to enhance performance on different 
tasks. Many works leverage a performance predictor, often called a surrogate model, to 
\emph{reliably} predict model accuracy. Such a surrogate can be trained through active learning
by querying a few models from the design space and regressing their performance to the remaining 
space (under some theoretical assumptions), 
thus significantly reducing search times \shortcite{nasbench_301,perf_predictors}. 

However, unlike CNN frameworks \shortcite{nasbench,efficientnet} that are only meant for vision tasks, there is no universal framework for NLP that 
differentiates among transformer architectural hyperparameters. Works that do 
compare different design decisions often do not consider \emph{heterogeneity} and \emph{flexibility} 
in their search space and explore the space over a limited hyperparameter set 
\shortcite{schubert,nas-bert,autobert_zero}\footnote{Here, by \emph{heterogeneity}, we mean that 
different encoder layers can have distinct attention operations, feed-forward stack depths, etc. By 
\emph{flexibility}, we mean that the hidden dimensions for different encoder layers, in a transformer architecture, are allowed to be 
mismatched.}. For instance, Primer~\shortcite{primer} only adds depth-wise convolutions to the attention 
heads; AutoBERT-Zero~\shortcite{autobert_zero} lacks deep feed-forward stacks; 
AutoTinyBERT~\shortcite{autotinybert} does not consider linear transforms (LTs) that have 
been shown to outperform 
traditional SA operations in terms of parameter efficiency; AdaBERT~\shortcite{adabert} 
only considers a design space of convolution and pooling operations. Most works, in the field of NAS for 
transformers, target model compression while trying to maintain the same performance~\shortcite{adabert,autotinybert,hat_mit}, which is 
orthogonal to our objectives in this work, \emph{i.e.}, searching for novel architectures that push 
the performance frontier. In addition, all previous works 
only consider \emph{rigid} architectures. For instance, DynaBERT~\shortcite{dynabert} only adapts the 
width of the network by varying the number of attention heads (and \emph{not} the hidden 
dimension of
each head), which is only a simple extension to traditional architectures. Further, their individual 
models still have the same hidden dimension throughout the network. 
AutoTinyBERT~\shortcite{autotinybert} and HAT~\shortcite{hat_mit}, among others, fix 
the input and output dimensions for each encoder layer (see
Appendix~\ref{app:self_attention} for a background on the SA
operation), which leads to \emph{rigid} architectures. 

\begin{table*}[]
\caption{Comparison of related works with different parameters (\checkmark indicates 
that the corresponding feature is present). Adaptive width refers to different architectures having 
possibly different hidden dimensions (albeit each layer within the architecture having the same hidden dimension). Full flexibility corresponds to 
each encoder layer having, possibly, a different hidden dimension.}
\small
\centering
\resizebox{\linewidth}{!}{  
\begin{tabular}{@{}l|cccccccccc@{}}
\toprule
\multirow{2}{*}{Framework} & \multicolumn{2}{c}{Self-Attention} & \multirow{2}{*}{Conv.} & \multicolumn{2}{c}{Lin. Transform} & \multirow{2}{*}{\begin{tabular}[c]{@{}c@{}}Flexible no. of \\ attn. ops.\end{tabular}} & \multirow{2}{*}{\begin{tabular}[c]{@{}c@{}}Flexible feed-\\ fwd. stacks\end{tabular}} & \multicolumn{2}{c}{Flexible hidden dim.} & \multirow{2}{*}{\begin{tabular}[c]{@{}c@{}}Search \\ technique\end{tabular}} \\ \cmidrule(lr){2-3} \cmidrule(lr){5-6} \cmidrule(lr){9-10}
 & SDP & WMA &  & DFT & DCT &  &  & Ad. width & Full flexibility &  \\ \midrule
Primer \\ \shortcite{primer} &  &  & \checkmark &  &  & \checkmark &  &  &  & ES \\
AdaBERT \\ \shortcite{adabert} &  &  & \checkmark &  &  &  &  &  &  & DS \\
AutoTinyBERT \\ \shortcite{autotinybert} & \checkmark &  &  &  &  & \checkmark & \checkmark & \checkmark &  & ST \\
DynaBERT \\ \shortcite{dynabert} & \checkmark &  &  &  &  & \checkmark & \checkmark & \checkmark &  & ST \\
NAS-BERT \\ \shortcite{nas-bert} & \checkmark &  & \checkmark &  &  & \checkmark &  &  &  & ST \\
AutoBERT-Zero \\ \shortcite{autobert_zero} & \checkmark &  & \checkmark &  &  & \checkmark &  &  &  & ES \\ \midrule
FlexiBERT (ours) & \checkmark & \checkmark & \checkmark & \checkmark & \checkmark & \checkmark & \checkmark & \checkmark & \checkmark & BOSHNAS \\ \bottomrule
\end{tabular}}
\label{tab:txf_nas_baselines}
\end{table*}

Table~\ref{tab:txf_nas_baselines} gives an 
overview of various baseline NAS frameworks for transformer architectures. It
presents the aforementioned works and the respective features they include.
Primer~\shortcite{primer} and AutoBERT-Zero~\shortcite{autobert_zero} exploit
evolutionary search (ES), which faces various drawbacks that limit \emph{elitist}
algorithms~\shortcite{elitist,bananas,nasbench_301}. AdaBERT~\shortcite{adabert}
leverages differentiable architecture search (DS), a popular technique used in many
CNN design spaces~\shortcite{nasbench_301}. On the other hand, some recent works like
AutoTinyBERT~\shortcite{autotinybert}, DynaBERT~\shortcite{dynabert}, and
NAS-BERT~\shortcite{nas-bert} leverage super-network training, where one large
transformer is trained and its sub-networks are searched in a \emph{one-shot} manner.
However, this technique is not amenable to diverse design spaces, as the
super-network size would drastically increase, limiting the gains from weight
transfer to the relatively minuscule sub-network. Moreover, previous works limit
their search to either the standard SA operation, \emph{i.e.}, the scaled
dot-product (SDP), or the convolution operation. We extend the basic attention
operation to also include the weighted multiplicative attention (WMA). Taking
motivation from recent advances with LT-based transformer models~\shortcite{fnet}, we also 
add discrete Fourier transform (DFT) and discrete cosine transform (DCT) to our design 
space. AutoTinyBERT and DynaBERT also allow adaptive widths in the transformer
architectures in their design space, however, each instance still has the same
dimensionality throughout the network (in other words, every encoder layer has the
same hidden dimension, as explained above). We detail why this is inherently a limitation 
in traditional transformer architectures in Appendix~\ref{app:self_attention}. FlexiBERT, to the best of our knowledge, is the first framework to allow full \emph{flexibility} -- not only different transformer instances in the design space can have distinct widths, but each encoder layer within a transformer instance can also have different hidden dimensions. Finally, we also leverage a novel NAS technique -- \underline{B}ayesian \underline{O}ptimization using \underline{S}econd-Order Gradients and \underline{H}eteroscedastic Models for \underline{N}eural \underline{A}rchitecture \underline{S}earch (BOSHNAS).

\subsection{Our contributions}

To address the limitations of \emph{homogeneous} and \emph{rigid} models, we make the following technical contributions:

\begin{itemize}
    \item We expand the design space of transformer hyperparameters to incorporate 
\textit{heterogeneous} architectures that venture beyond simple SA by employing other 
operations like convolutions and LTs.
    \item We propose novel projection layers and \emph{relative}/\emph{trained} positional encodings to make hidden sizes \emph{flexible} across layers -- hence the name FlexiBERT. 
    \item We propose \texttt{Transformer2vec} that uses similarity measures to compare
computational graphs of transformer models to obtain a dense embedding that captures model similarity in a Euclidean space.
    \item We propose a novel NAS framework, namely, BOSHNAS. It uses a 
neural network as a heteroscedastic surrogate model and second-order gradient-based optimization using 
backpropagation to input (GOBI)~\shortcite{tuli2021cosco} to speed up search for the 
next query in the exploration process. It leverages \emph{nearby} trained models to transfer 
weights in order to reduce the amortized search time for every query.
    \item Experiments on the GLUE benchmark \shortcite{glue} show that BOSHNAS applied to the 
FlexiBERT design space results in a score improvement of \perf compared to the baseline, \emph{i.e.}, NAS-BERT~\shortcite{nas-bert}. The proposed model, FlexiBERT-Mini, has \parambertmini fewer parameters than BERT-Mini and achieves \perfbertmini higher GLUE score. FlexiBERT also outperforms the best \emph{homogeneous} architecture by 3\%, while requiring \paramhomo fewer parameters. FlexiBERT-Large, our BERT-Large~\shortcite{bert} counterpart, outperforms the state-of-the-art models by at least \largeperf average accuracy on the first eight tasks in the GLUE benchmark~\shortcite{glue}.
\end{itemize}

The rest of the paper is organized as follows. Section~\ref{sec:related_work} presents related work. Section~\ref{sec:methods} describes the set of steps and decisions that undergird the \mbox{FlexiBERT} framework. In Section~\ref{sec:results}, we present the results of design space exploration experiments. Finally, Section~\ref{sec:conclusion} concludes the article.

\section{Related Work}
\label{sec:related_work}

We briefly describe related work next.

\subsection{Transformer design space}

Traditionally, transformers have primarily relied on the SA operation (details in Appendix~\ref{app:self_attention}). Nevertheless, several works have proposed various compute blocks to reduce the number of model parameters and hence computational cost, without compromising performance. For instance, ConvBERT uses dynamic 
span-based convolutional operations that replace SA heads to directly model local dependencies \shortcite{convbert}. Recently, FNet improved model efficiency using 
LTs instead \shortcite{fnet}. MobileBERT, another recent architecture, uses bottleneck structures and multiple feed-forward stacks to obtain smaller 
and faster models while achieving competitive results on well-known benchmarks~\shortcite{mobilebert}. For completeness, we present other previously proposed advances to improve the BERT model in Appendix~\ref{app:bert_improv}.

\subsection{Neural architecture search}

NAS is an important machine learning technique that algorithmically searches for new
neural network architectures within a pre-specified design space under a given objective~\shortcite{automl_survey}. Prior work has implemented NAS using a variety of techniques, albeit limited to the CNN design 
space. A popular approach is to use a reinforcement learning algorithm, REINFORCE, that has 
been shown to be superior to other tabular approaches \shortcite{reinforce}. Other approaches 
include Gaussian-Process-based Bayesian Optimization (GP-BO) \shortcite{gp_bo}, Evolutionary Search (ES) \shortcite{evolutionary_search,nsga}, etc. However, these methods come with challenges that limit their ability to reach state-of-the-art results in the CNN design space~\shortcite{bananas}. 

Recently, NAS has also seen application of surrogate models for performance prediction in 
CNNs \shortcite{nasbench_301}. This results in training of much fewer models to predict accuracy for the entire design space under some confidence constraints. However, these predictors are computationally expensive to train. This leads to a bottleneck, especially in large design spaces, in the training of subsequent models since new queries are produced only after this predictor is trained for every batch of trained models in the search space. \shortciteA{nasbench_301} use a Graph Isomorphism Net \shortcite{gin} that regresses performance values directly on the computational graphs formed for each CNN model.

Although previously restricted to CNNs \shortcite{nasnet}, NAS has recently seen applications 
in the transformer space as well. \shortciteA{evolved_txf} use standard NAS techniques to search 
for optimal transformer architectures. However, their method requires every new model to be 
trained from scratch. They do not employ knowledge transfer, in which weights from previously
trained \emph{neighboring} models are used to speed up subsequent training. This is important in
the transformer space since pre-training every model is computationally expensive. Further, the
attention heads in their model follow the same dimensionality, \emph{i.e.}, are not fully \emph{flexible}.

One of the state-of-the-art NAS techniques, BANANAS, implements Bayesian Optimization (BO) 
over a neural network model and predicts performance uncertainty using ensemble networks 
that are, however, too compute-heavy \shortcite{bananas}. BANANAS uses mutation/crossover on the current set of best-performing models and obtains the next best predicted model in this local 
space. Instead, we propose the use of GOBI~\shortcite{tuli2021cosco} in order to efficiently search for the next query in the \emph{global} space. Thanks to random cold restarts, GOBI can search over diverse 
models in the architecture space. BANANAS also uses path embeddings, which have been shown to 
perform sub-optimally for search over a diverse space \shortcite{nasgem}. 

\subsection{Graph Embeddings that Drive NAS}
\label{sec:nasgem_wl}

Many works on NAS for CNNs have primarily used graph embeddings to model their performance predictor. These embeddings are formed for each \emph{computational graph}, representing a specific CNN architecture in the design space. A popular approach to learning with graph-structured data is to make use of graph 
kernel functions that measure similarity between graphs. 
A recent work, NASGEM \shortcite{nasgem}, uses the Weisfeiler-Lehman (WL) sub-tree kernel, which compares tree-like 
substructures of two computational graphs. This helps distinguish between substructures that other kernels, 
like the random walk kernel, may deem identical \shortcite{wl_kernel}. Also, the WL kernel has an 
attractive computational complexity. This has made it one of the most widely used graph kernels. 
Graph-distance-driven NAS often leads to enhanced representation capacity that yields optimal 
search results \shortcite{nasgem}.
However, the WL kernel only computes sub-graph similarities based on overlap in graph nodes. 
It does not consider whether two nodes are \emph{inherently} similar or not. For example, a 
computational `block' (or its respective graph node) for an SA head with $h=128$ and 
$o=$ SDP, would be closer to another attention block with, say, $h=256$ and $o=$ WMA, but would be 
farther from a block representing a feed-forward layer (for details on SA types, 
see Section~\ref{sec:des_space}).

Once we have similarities computed between every possible graph pair in the
design space, next we learn dense embeddings, the Euclidean distance for which should follow the
similarity function. These embeddings would not only be helpful in effective visualization of the
design space, but also for fast computation of \emph{neighboring} graphs in the active-learning
loop. Further, a dense embedding helps us practically train a finite-input surrogate function (as 
opposed to the sparse path encodings used in \shortciteR{bananas}).
Many works have achieved this using different techniques. \shortciteA{graph2vec} 
train task-specific graph embeddings using a skip-gram model and negative sampling, taking inspiration 
from \texttt{word2vec} \shortcite{word2vec}. In this work, we take inspiration from \texttt{GloVe} 
instead \shortcite{glove}, by applying manifold learning to all distance pairs 
\shortcite{kruskal_mds_1964}. Hence, using global similarity distances built over domain knowledge, 
and batched gradient-based training, we obtain the proposed \texttt{Transformer2vec} embeddings that 
are superior to traditional generalized graph embeddings.

We take motivation from NASGEM \shortcite{nasgem}, which showed that training a WL kernel-guided encoder has advantages in scalable and flexible search. Thus, we train a performance predictor on the \texttt{Transformer2vec} embeddings, which not only aid in transfer of weights between neighboring models, but also support better-posed continuous performance approximation. % We also test the hypothesis that similar models have similar performance for transformers, which has been shown for CNNs \shortcite{nasbench}. 
More details on the computation of these embedding are given in Section~\ref{sec:transformer2vec}.

\section{Methodology}
\label{sec:methods}

\begin{figure}
\setlength{\floatsep}{-20pt}
    \centering 
    \includegraphics[width=0.6\linewidth]{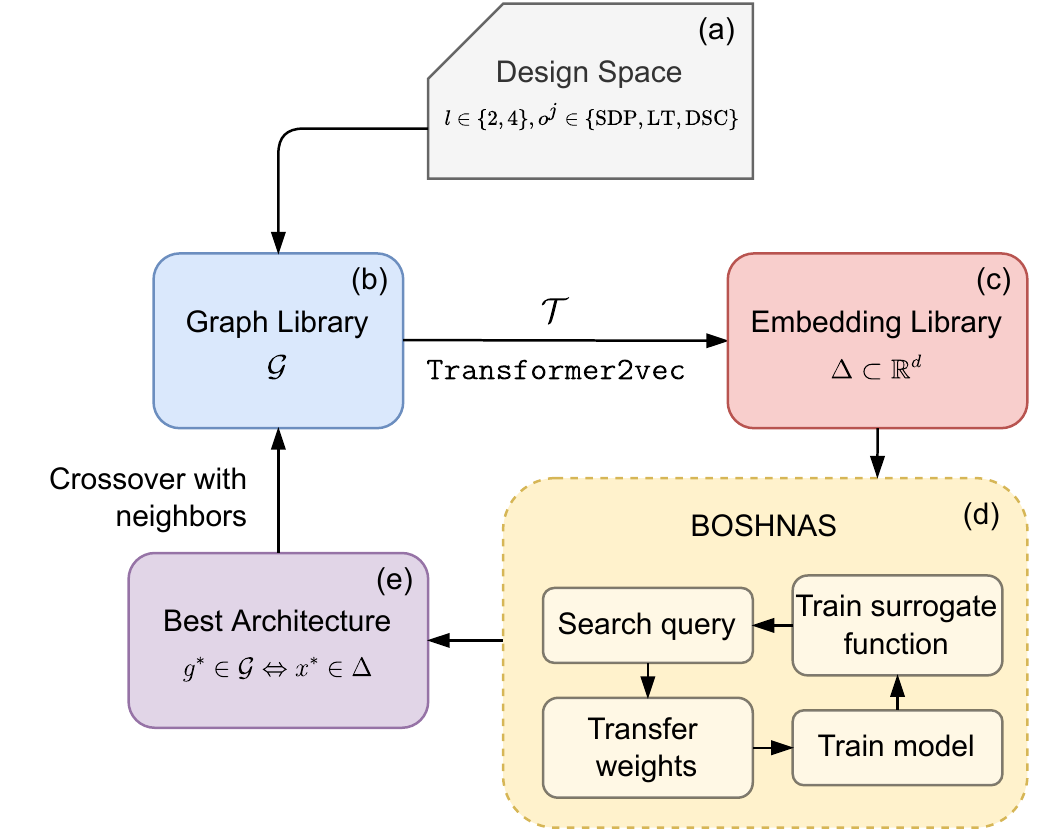}
    \caption{Overview of the FlexiBERT pipeline.}
    \label{fig:flowchart}
\end{figure}

In this work, we train a heteroscedastic surrogate model that predicts the performance of a transformer architectrue and uses it to run second-order optimization in the design space. We do this by decoupling the 
training procedure from pre-processing of the embedding of every model in the design space to speed up training. 
First, we train embeddings to map the space of computational graphs to a Euclidean space (\texttt{Transformer2vec})
and then train the surrogate model on the embeddings.

% Figure~\ref{fig:flowchart} summarizes the complete pipeline where we iteratively train an
% embedding on the computational graphs ($\mathcal{G}_i$) in the FlexiBERT design space. Here, $i$ denotes the iteration index, which we drop in subsequent discussion when unambiguous. 

Our work involves exploring a vast and heterogeneous design space and searching for optimal
architectures with a given task. To this end, we (a) define a design space via a flexible
set of architectural choices (see Section~\ref{sec:des_space}), (b) generate possible
computational graphs ($\mathcal{G}$; see Section~\ref{sec:comp_graph}), (c) learn an
embedding for each point in the space using a distance metric for graphs ($\Delta$; see
Section~\ref{sec:transformer2vec}), and (d) employ a novel search technique (BOSHNAS) based
on surrogate modeling of the performance and its uncertainty over the continuous embedding
space (see Section~\ref{sec:boshnas}). In addition, to tackle the enormous design space, we
propose a hierarchical search technique that iteratively searches over finer-grained
models derived from (e) a crossover of the best models obtained in the current iteration
and their neighbors. Figure~\ref{fig:flowchart} gives a broad overview of the FlexiBERT pipeline as explained above. An unrolled version of this iterative flow is presented below:
\begin{equation*}
    \text{Design Space} \rightarrow \mathcal{G}_1 \xrightarrow{\mathcal{T}} \Delta_1 \xrightarrow{\text{BOSHNAS}} g^* \xrightarrow{\text{cross-over}} \mathcal{G}_2 \xrightarrow{\mathcal{T}} \ldots
\end{equation*}
However, for simplicity of notation, we omit the iteration index in further references. We now discuss the key elements of this pipeline in detail.

\subsection{FlexiBERT Design Space}
\label{sec:des_space}

We now describe the FlexiBERT design space, \textit{i.e.}, box (a) in Figure~\ref{fig:flowchart}.

\subsubsection{Set of operations in FlexiBERT}

The traditional BERT model is composed of multiple layers, each containing a bidirectional 
multi-headed by SA module followed by a feed-forward module. Several 
modifications have been proposed to the original encoder, primarily to the attention
module. This gives rise to a richer design space. We consider WMA-based SA in addition to SDP-based 
operations \shortcite{wma}. 

We also incorporate LT-based attention in FNet~\shortcite{fnet}
and dynamic-span-based convolution (DSC) in ConvBERT~\shortcite{convbert}, in place
of the vanilla SA mechanism. Whereas the original FNet 
implementation uses DFT, we also consider DCT. The motivation behind using DCT is its widespread application 
in lossy data compression, which we believe can lead to sparse weights, thus leaving room for 
optimizations with sparsity-aware machine learning accelerators~\shortcite{spring}. Our design space 
allows variable kernel sizes for convolution-based 
attention.  Consolidating different attention module types that vary in their computational 
costs into a single design space enables the models to have inter-layer variance in 
expression capacity. Inspired by MobileBERT~\shortcite{mobilebert}, we also consider architectures with multiple feed-forward stacks. We summarize the entire design space with the range of each operation type in
Table~\ref{tab:hyp_ranges}.  The ranges of different hyperparameters are in accordance 
with the design space spanned by BERT-Tiny to BERT-Mini \shortcite{turc2019}, with additional 
modules included as discussed. We call this the Tiny-to-Mini space. This restricts our curated testbed to models with up to $3.3 \times 10^{7}$ trainable parameters. This curated parameter space allows us to perform extensive experiments, comparing the proposed approach against various baselines (see Section~\ref{sec:results_boshnas_ablation}).

\begin{table}[t]
\caption{Design space description. Super-script ($j$) depicts the value for layer $j$.}
\vskip 0.1in
    \centering
    \resizebox{0.7\linewidth}{!}{  
    \begin{tabular}{@{}ll@{}}
    \toprule
        \textbf{Design Element} & \textbf{Allowed Values} \\
        \midrule 
        Number of encoder layers ($l$) & $\{2,4\}$\\
        Type of attention operation used ($o^j$) & \{SA, LT, DSC\}\\
        Number of operation heads ($n^j$) & $\{2,4\}$ \\
        Hidden size ($h^j$) & $\{128,256\}$\\
        Feed-forward dimension ($f^j$) & $\{512,1024\}$\\
        Number of feed-forward stacks & $\{1,3\}$\\
        Operation parameters ($p^j$): & \\
       \hspace{3mm} if $o^j =$ SA & Self-attention type: \{SDP, WMA\} \\ 
       \hspace{3mm} else if $o^j =$ LT & Linear transform type: \{DFT \;, DCT\} \\
       \hspace{3mm} else if $o^j =$ DSC & Convolution kernel size: $\{5,9\}$ \tabularnewline
       \bottomrule
    \end{tabular}}
    \label{tab:hyp_ranges}
\end{table}

Every model in the design space can therefore be expressed via a model card that is a 
dictionary containing the chosen value for each design decision. BERT-Tiny~\shortcite{turc2019}, in this formulation, can be represented as
\begin{align*}
    \Big\{ l: 2, o: [\text{SA}, \text{SA}], h: [128, 128], n: [2, 2],
    f: \left[[512], [512]\right], p: [\text{SDP}, \text{SDP}] \Big\}.
\end{align*}
where the length of the list for every entry in $f$ denotes the size of the feed-forward stack. The model card can be used to derive the computational graph of the model using smaller 
modules inferred from the design choice (details in Section~\ref{sec:comp_graph}).

\subsubsection{Flexible hidden dimensions}

In traditional transformer architectures, the flow of information is restricted through
the use of a constant embedding dimension across the network (a matrix of dimensions $N_T \times h$ from one layer to the next, where $N_T$ denotes the number of tokens and $h$ the hidden dimension; more details in Appendix~\ref{app:self_attention}). We allow architectures in our design space to have \emph{flexible} dimensions across layers that, in turn, enables different layers to capture information of different dimensions, as it learns more abstract features deeper into the network. For this, we make the following modifications:

\begin{itemize}
    \item \textit{Projection layers}: We add an affine projection network between encoder layers 
with dissimilar hidden sizes to transform encoding dimensionality. 
    \item \textit{Relative positional encoding}: The vanilla-BERT implementation uses an 
absolute positional encoding at the input and propagates it ahead through residual connections. 
Since we relax the restriction of a constant hidden size across layers, this is not applicable 
to many models in our design space (as the learned projections for absolute encodings may not be one-to-one). Instead, we add a \emph{relative} positional encoding 
at each layer \shortcite{shaw2018selfattention,huang2018music,xlnet}. Such an encoding can entirely replace absolute 
positional encodings with relative position representations learned using the 
SA mechanism. Whereas the SA module implementation remains the same 
as in previous works, for DSC-based and LT-based attention, we learn 
the relative encodings separately using SA and add them to the output of
the attention module. 

Formally, let $\mathbf{Q}$ and $\mathbf{V}$ denote the query and the value layers, 
respectively. Let $\mathbf{R}$ denote the relative embedding tensor that is to be learned. 
Let $\mathbf{Z}$ and $\mathbf{X}$ denote the output and the input tensors of the attention 
module, respectively. In addition, let us define LT-based attention and 
DSC-based attention as $\text{LT}(\cdot)$ and $\text{DSC}(\cdot)$, respectively. Then,
    \begin{equation*}
        \mathrm{RelativeAttention}(\mathbf{X}) = \mathrm{softmax} \left(\frac{\mathbf{QR}^\top}{\sqrt{d_{\mathbf{Q}}}}\right)\mathbf{V}
    \end{equation*}
    \begin{align*}
        \mathbf{Z}_{\text{LT}} &= \text{LT}(\mathbf{X}) + \mathrm{RelativeAttention}(\mathbf{X}) \\
        \mathbf{Z}_{\text{DSC}} &= \text{DSC}(\mathbf{X}) + \mathrm{RelativeAttention}(\mathbf{X})
    \end{align*}
\end{itemize}

It should be noted that the proposed approach would only be applicable when the 
positional encodings are \emph{trained}, instead of being predetermined \shortcite{vaswani}. 
Thanks to \emph{relative} and \emph{trained} positional encodings, this enabled us to make the 
dimensionality of data flow \emph{flexible} across the network layers. This also means that each layer 
in the feed-forward stack can have a distinct hidden dimension.

\subsection{Graph Library}
\label{sec:comp_graph}

We now describe the graph library, \textit{i.e.}, box (b) in Figure~\ref{fig:flowchart}.

\subsubsection{Block-level computational graphs}

To learn a lower-dimensional dense manifold of the given design space, characterized by a 
large number of FlexiBERT models, we convert each model into a computational graph. This is formulated based on the forward flow of connections for 
each compute block. For our design space, we take all possible combinations of the compute blocks derived from the design decisions presented in Table~\ref{tab:hyp_ranges} (see Appendix~\ref{app:compute_blocks}).

Using this design space and the set of compute blocks, we create all possible computational graphs within the design space for every
transformer model. We then
use recursive hashing as follows~\shortcite{nasbench}. For every node in this graph, we 
concatenate the hash of its input, hash of that node, and its output, and then take the hash 
of the result. We use SHA256 as our hashing function. Doing this for all nodes and then hashing the concatenated hashes gives us the resultant hash of a given computational graph. 
This helps us detect isomorphic graphs and remove redundancy.

\begin{figure}[t]
    \centering %\setlength{\belowcaptionskip}{-15pt}
    \includegraphics[width=0.6\linewidth]{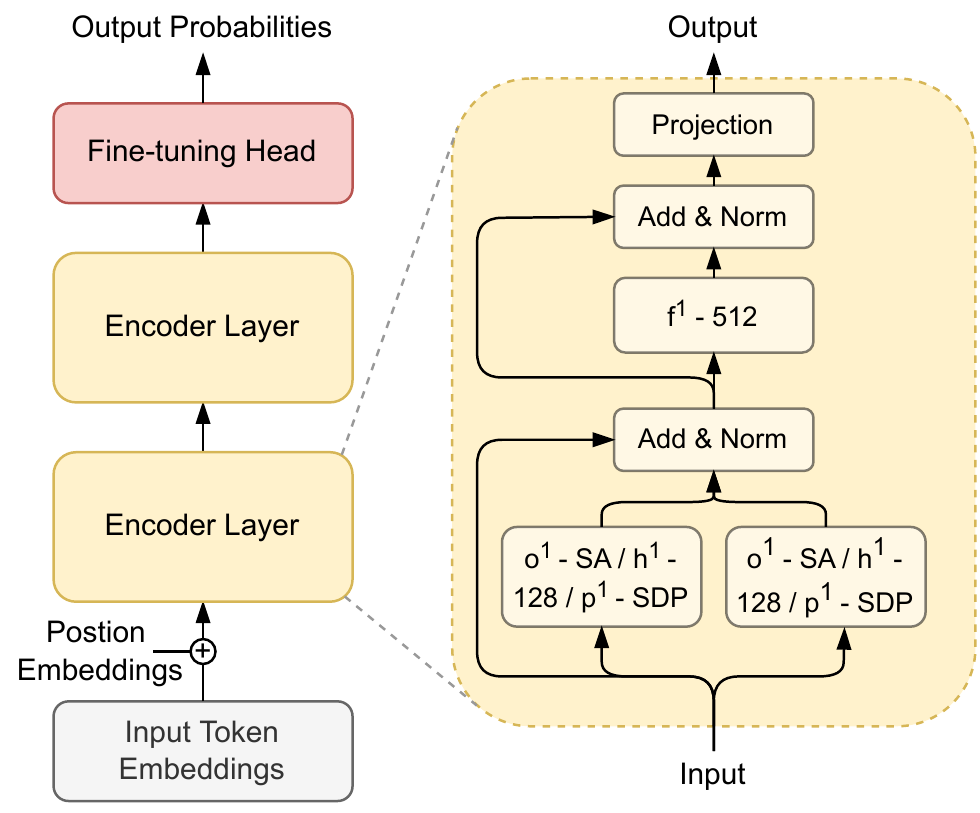}
    \caption{Block-level computation graph for BERT-Tiny in FlexiBERT. The projection 
layer implements an identity function since the hidden sizes of input and output encoder layers are 
equal.}
    \label{fig:bert_tiny_comp_graph}
\end{figure}

Figure~\ref{fig:bert_tiny_comp_graph} shows the block-level computational graph for BERT-Tiny. Using the connection patterns for every possible block permutation (as presented in Appendix~\ref{app:compute_blocks}), we can generate 
multiple graphs for the given design space.

\subsubsection{Levels of hierarchy}

The total number of possible graphs in 
the design space with \emph{heterogeneous} feed-forward hidden layers 
is $\sim$3.32 billion. This is substantially larger than any transformer design space used in the past.

To make our approach tractable, we propose a hierarchical search method. 
Each model in the design space can be considered to be composed of multiple stacks, where
each stack contains at least one encoder layer. In the first step, we restrict each stack
to $s=2$ layers, where each layer in a stack shares the same design configuration.
Naturally, this limits the search space size (the set of all graphs in this space is denoted by
$\mathcal{G}_1$). Hence, for instance, BERT-Tiny falls under $\mathcal{G}_1$ since the two encoder 
layers have the same configuration.  We learn embeddings in this space
and then run NAS to obtain the best-performing models. In the
subsequent step, we consider a design space constituted by a finer-grained neighborhood of
these models. The neighborhood is derived by pairwise crossover between the best-performing 
models and their neighbors in a space where the number of layers per stack
is set to $s/2=1$, denoted by $\mathcal{G}_2$ (details in Appendix~\ref{app:crossover}). Finally, we include \emph{heterogeneous} feed-forward stacks ($s=1^*$) and denote the space by $\mathcal{G}_3$.

\subsection{\texttt{Transformer2vec}}
\label{sec:transformer2vec}

We now describe the \texttt{Transformer2vec} embedding and how we create
an embedding library
from a graph library $\mathcal{G}$, \textit{i.e.}, box (c) in Figure~\ref{fig:flowchart}.

\subsubsection{Graph edit distance}

Taking inspiration from \shortciteA{nasgem} and \shortciteA{glove}, we train dense embeddings using \emph{global} distance metrics, such as the Graph Edit Distance (GED)~\shortcite{ged_networkx}. 
These embeddings enable fast derivation of 
\emph{neighboring} graphs in the active learning loop to facilitate transfer of weights. 
We call them \texttt{Transformer2vec} embeddings. Unlike other approaches like the 
WL kernel, GED bakes in \emph{domain knowledge} in graph comparisons, as explained in Section~\ref{sec:nasgem_wl}, by using a weighted sum of node insertions,
deletions, and substitutions costs. %Hence, we use GED in our work.

For the GED computation, we first sort all possible compute blocks in the order of their 
computational complexity. Then, we weight the insertion and deletion cost for every block based 
on its index in this sorted list, and the substitution cost between two blocks based on the 
difference in the indices in this sorted list. For computing the GED, we use a depth-first 
algorithm that requires less memory than traditional methods \shortcite{ged_networkx}.

\subsubsection{Training embeddings} 

Given that there are $S$ graphs in $\mathcal{G}$, we compute the GED for all possible computational graph pairs. This gives us a dataset of $N = {S \choose 2}$ distances. To train the embedding, we minimize the mean-square error as the loss function between the predicted Euclidean 
distance and the corresponding GED. For the design space in consideration, embeddings of 
$d$ dimensions are generated for every level of the hierarchy. Concretely, to train embedding $\mathcal{T}$, we minimize the loss
\begin{align*}
    \mathcal{L}_{\mathcal{T}} = \sum_{1 \le i \le N, 1 \le j \le N, i \ne j} \ \Big( \mathbf{d}(\mathcal{T}({g_i}), \mathcal{T}({g_j})) - \mathrm{GED}(g_i, g_j) \Big)^2,
\end{align*}
where $\mathbf{d}(\cdot, \cdot)$ is the Euclidean distance and 
the $\mathrm{GED}$ is calculated for the corresponding computational graphs 
$g_i$, $g_j \in \mathcal{G}$.

\subsubsection{Weight transfer among neighboring models}

Pre-training each model in the design space is computationally expensive. Hence, we rely on weight 
sharing to initialize a query model in order to \emph{directly} fine-tune it and minimize exploration time (details in Appendix~\ref{app:know_transfer}). 
We do this by generating $k$ nearest neighbors of a graph in the design space (we use $k=100$ for our 
experiments).  Naturally, we would 
like to transfer weights from the corresponding fine-tuned neighbor that is closest to the query, 
as such models intuitively have similar initial internal representations.

We calculate this similarity using a \textit{biased overlap} measure that counts the number 
of encoder layers from the input to the output that are common to the current graph 
(\emph{i.e.}, have exactly the same set of hyperparameter values). We stop counting the overlap when 
we encounter different encoder layers, regardless of subsequent overlaps. In this ranking, there could be more than one graph with the same biased 
overlap with the current graph. Since the internal representations learned depend on 
the subsequent set of operations as well, we break ties based on the embedding distance of 
these graphs with the current graph. This gives us a set of neighbors, denoted by $N_q$ for a model $q$, for every graph that 
are ranked based on both the biased overlap and the embedding distance. It helps 
increase the probability of finding a trained neighbor with high overlap.

As a hard constraint, we only consider transferring weights if the \emph{biased overlap 
fraction} ($\mathcal{O}_f(q, n) = \text{\emph{biased overlap}}/l_q$, where $q$ is the query 
model, $n \in N_q$ is the neighbor in consideration, and $l_q$ is the number of layers in $q$) 
between the queried model and its neighbor is above a threshold $\tau$. If the constraint is met, the weights of the shared part from the 
corresponding neighbor are transferred to the query and the query is fine-tuned. Otherwise, we pre-train the query. The weight transfer operation is denoted by $W_q \gets W_n$.

\subsection{BOSHNAS}
\label{sec:boshnas}

We now describe the BOSHNAS search policy, \textit{i.e.}, box (d) in Figure~\ref{fig:flowchart}.

\subsubsection{Uncertainty types} 

To overcome the challenges of an unexplored design space, it is important to consider the 
uncertainty in model predictions to guide the search process. Predicting model performance 
deterministically is not enough to estimate the next most probably best-performing model. 
We leverage the upper confidence bound (UCB)  exploration on the predicted performance of unexplored models~\shortcite{russel2010}. This could arise from not only the approximations in the surrogate modeling 
process, but also parameter initializations and variations in model performance due to different 
training recipes. These are called \emph{epistemic} and \emph{aleatoric} uncertainties, respectively. The former, also called reducible uncertainty, arises from a lack of knowledge or information, and the latter, also called irreducible uncertainty, refers to the inherent variation in the 
system to be modeled.

\subsubsection{Surrogate model}  

In BOSHNAS, we use Monte-Carlo (MC) dropout \shortcite{mc_dropout} and a Natural Parameter Network (NPN)~\shortcite{npn} to model the epistemic and aleatoric uncertainties, respectively. 
The NPN not only helps with a distinct prediction of aleatoric uncertainty that can be used for 
optimizing the training recipe once we are close to the optimal architecture, but also serves 
as a superior model than Gaussian Processes, Bayesian Neural Networks (BNNs), and other 
Fully-Connected Neural Networks (FCNNs)~\shortcite{tuli2021cosco}.  Consider the NPN network $f_S(x; \theta)$ with a transformer embedding 
$x$ as an input and parameters $\theta$. The output of such a network is the pair 
$(\mu, \sigma) \gets f_S(x; \theta)$, where $\mu$ is the predicted mean performance and 
$\sigma$ is the aleatoric uncertainty.  To model the epistemic uncertainty, we use two deep 
surrogate models: (1) teacher ($g_S$) and (2) student ($h_S$) networks. It is a 
surrogate for the performance of a transformer, using its embedding $x$ as an input. The 
teacher network is an FCNN with MC Dropout (parameters $\theta'$). To compute the epistemic 
uncertainty, we generate $n$ samples using $g_S(x, \theta')$. The standard deviation of the 
sample set is denoted by $\xi$. To run GOBI \shortcite{tuli2021cosco} and avoid numerical gradients 
due to their poor performance, we use a student network (FCNN with parameters $\theta''$) that 
directly predicts the output $\hat{\xi} \gets h_S(x, \theta'')$, a surrogate of $\xi$. 

\subsubsection{Active learning and optimization} 

For a design space $\mathcal{G}$, we first form an embedding space $\Delta$ by transforming all graphs in $\mathcal{G}$ using the \texttt{Transformer2vec} embedding. Assuming we have the three networks $f_S, g_S$, and $h_S$, for our surrogate model, 
we use the following UCB estimate:
\begin{flalign}
\label{eq:ucb}
\begin{split}
    \mathrm{UCB} &= \mu + k_1 \cdot \sigma + k_2 \cdot \hat{\xi} = \Big(f_S(x, \theta)[0] + k_1 \cdot f_S(x; \theta)[1]\Big) + k_2 \cdot h_S(x, \theta''),
\end{split}
\end{flalign}
where $x \in \Delta$, $k_1$, and $k_2$ are hyperparameters. 

To generate the next transformer to test, we run 
GOBI using neural network inversion and the AdaHessian optimizer~\shortcite{yao2021adahessian} that uses second-order updates 
to $x$ ($\nabla^2_x \mathrm{UCB}$) up till convergence. From this, we get a new query embedding, 
$x'$. The nearest transformer architecture is found based on the Euclidean distance of all 
available transformer architectures in the design space $\Delta$, giving the next closest model 
$x$. This model is fine-tuned (or pre-trained if there is no nearby trained model with sufficient 
overlap; see Section~\ref{sec:transformer2vec}) on the required task to give the respective 
performance. Once the new datapoint, $(x, o)$, is received, we train the models using the loss 
functions on the updated corpus, $\delta'$:
\begin{equation}
\label{eq:boshnas_losses}
\begin{aligned}
\mathcal{L}\textsubscript{NPN}(f_S, x, o) &= \sum_{(x, o) \in \delta'} \frac{(\mu - o)^2}{2 \sigma^2} + \frac{1}{2}\ln{\sigma^2},\\
\mathcal{L}\textsubscript{Teacher}(g_S, x, o) &= \sum_{(x, o) \in \delta'} (g_S(x, \theta') - o)^2,\\
\mathcal{L}\textsubscript{Student}(h_S, x) &= \sum_{x, \forall (x, o) \in \delta'} (h_S(x, \theta'') - \xi)^2,
\end{aligned}
\end{equation}
where $\mu,\sigma $ = $ f_S(x, \theta)$, and $\xi$ is obtained by sampling $g_S(x, \theta')$. The first is 
the aleatoric loss to train the NPN model \shortcite{npn}; the other two are squared-error loss 
functions. Appendix~\ref{app:boshnas_flow} presents the flow of these models in a schematic. We run multiple random cold restarts of GOBI to get multiple queries for the next 
step in the search process.

\SetKwComment{Comment}{/* }{ */}
\begin{algorithm}[t]
\DontPrintSemicolon
\SetAlgoLined
\KwResult{\textbf{best} architecture}
\textbf{Initialize:} overlap threshold ($\tau$), convergence criterion, uncertainty sampling 
prob. ($\alpha$), diversity sampling prob. ($\beta$), surrogate model ($f_S$, $g_S$, and 
$h_S$) on initial corpus $\delta$, design space $g \in \mathcal{G} \Leftrightarrow x \in \Delta$; \par
\While{convergence criterion not met}{
  wait till a worker is free\;
  \eIf{prob $\sim U(0,1) < 1 - \alpha - \beta$}{
   $\delta \gets \delta \ \cup$ \{new performance point ($x, o$)\}; \par
   fit(\textbf{surrogate}, $\delta$) using Eqn.~\eqref{eq:boshnas_losses}; \par 
   \label{line:fit} $x$ $\gets$ GOBI($f_S$, $h_S$); \Comment*{Optimization step} \label{line:opt}
    \For{$n$ in $N_x$}{
     \If{$n$ is trained \& $\mathcal{O}_f(x, n) \ge \tau$}{
      $W_x \gets W_n$; \par
      send $x$ to worker; \label{line:train} \par
      $\textbf{break}$;
     }
    }
  }{
   \eIf{$1 - \alpha - \beta \le$ prob. $< 1 - \beta$}{
    $x$ $\gets$ $\textbf{argmax}$($k_1 \cdot \sigma + k_2 \cdot \hat{\xi}$); \label{line:uncertainty} \Comment*{Uncertainty sampling}
    send $x$ to worker; 
   }{
    send random $x$ to worker; \Comment*{Diversity sampling} \label{line:diversity}
   }
  }
 }
 \caption{BOSHNAS} 
 \label{alg:boshnas}
\end{algorithm}

Algorithm~\ref{alg:boshnas} summarizes the BOSHNAS workflow. Starting from an initial pre-trained set $\delta$ in the first level of hierarchy
$\mathcal{G}_1$, we run until convergence the following steps in a multi-worker compute
cluster. To trade off between exploration and exploitation, we consider two probabilities:
uncertainty-based exploration ($\alpha$) and diversity-based exploration ($\beta$). With
probability $1 - \alpha - \beta$, we run second-order GOBI using the surrogate
model to minimize UCB in Eq.~\eqref{eq:ucb}. Adding the converged point $(x, o)$ in
$\delta$, we minimize the loss values in Eq.~\eqref{eq:boshnas_losses}
(line~\ref{line:fit} in Algorithm~\ref{alg:boshnas}). We then generate a new query point,
transfer weights from a neighboring model, and train it
(lines~\ref{line:opt}-\ref{line:train}). With $\alpha$ probability, we sample the search
space using the combination of aleatoric and epistemic uncertainties, 
$k_1 \cdot \sigma + k_2 \cdot \hat{\xi}$, to find a point where the performance estimate is uncertain
(line~\ref{line:uncertainty}). To avoid getting stuck in a localized search subset, we also
choose a random point with probability $\beta$ (line~\ref{line:diversity}). Once we converge
in the first level, we continue with second and third levels, $\mathcal{G}_2$ and
$\mathcal{G}_3$, as described in Section~\ref{sec:comp_graph}.

\section{Experimental Results}
\label{sec:results}

In this section, we show how the FlexiBERT model obtained from BOSHNAS outperforms the baselines.

\subsection{Setup}
For our experiments, we set the number of layers in each stack to $s=2$ for the first level of the 
hierarchy, where models have the  same configurations in every stack. In the second
level, we use $s=1$. Finally, we also make the feed-forward stacks heterogeneous ($s=1^*$) in the 
third level (details given in Section~\ref{sec:comp_graph}). For the range of design choices in 
Table~\ref{tab:hyp_ranges} and setting $s=2$, we obtained 9312 unique graphs after removing 
isomorphic graphs. The dimension of the \texttt{Transformer2vec} embedding is set to $d=16$ after running grid search. 
To do this, we  minimize the distance prediction error while also keeping $d$ small using knee-point detection.
The hyperparameter values in Algorithm~\ref{alg:boshnas} are obtained through grid search. We use 
overlap threshold  $\tau=80\%$, $\alpha = \beta = 0.1$, and $k_1 = k_2 = 0.5$ in our experiments.  
The convergence criterion is met in BOSHNAS when the change in performance is within $10^{-4}$
for five iterations. Further experimental setup details are given in Appendix~\ref{app:model_training}.

\begin{figure}[t]
    \centering \setlength{\belowcaptionskip}{0pt}
    \includegraphics[width=0.6\linewidth]{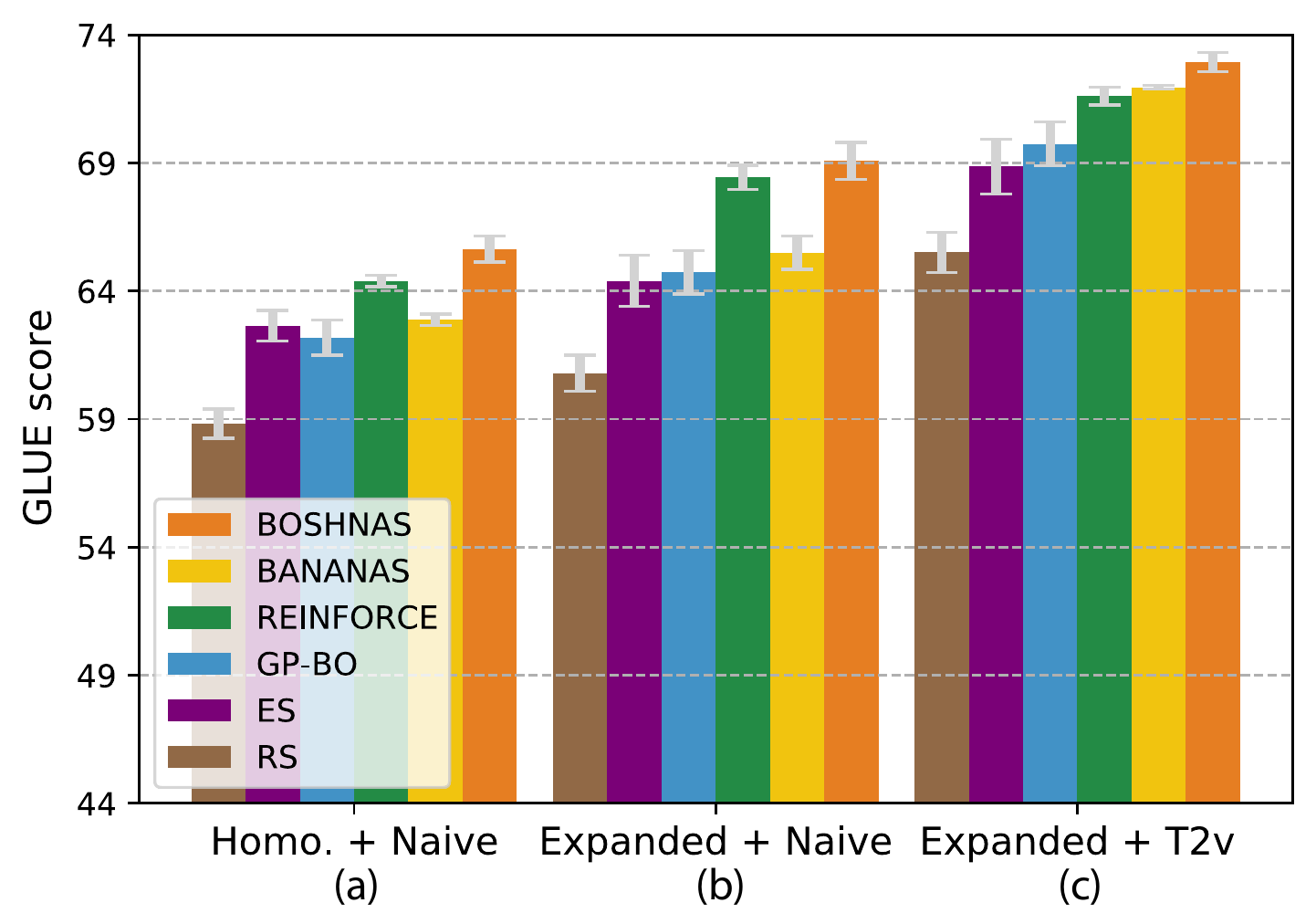}
    \caption{Bar plot comparing all NAS techniques with (a) naive 
embeddings and a design space of homogeneous models, (b) naive embeddings and an expanded design space of 
homogeneous and heterogeneous models, and (c) \texttt{Transformer2vec} (T2v) embeddings with 
the expanded design space. Plotted with 90\% confidence intervals.}
    \label{fig:ablation_baselines}
\end{figure}

\subsection{Pre-training and Fine-tuning Models}
\label{sec:pretraining_finetuning}

Our pre-training recipe is adapted from the one used in RoBERTa, proposed by \shortciteA{roberta}, with slight
variations in order to reduce the training budget (details in Appendix~\ref{app:model_training}).

We initialize the architecture space with models adapted from equivalent models presented in 
literature~\shortcite{turc2019,fnet,convbert}. The
12 initial models used to initiate the search process are BERT-Tiny, BERT-2/256 (with two encoder 
layers and a fixed hidden dimension of 256), BERT-4/128, BERT-Mini, FNet-Tiny, FNet-2/256, FNet-4/128, 
FNet-Mini, ConvBERT-Tiny, ConvBERT-2/256, ConvBERT-4/128, and ConvBERT-Mini (with  $p^j =$ DFT for 
FNets and $p^j = 9$ for ConvBERTs adapted from the original models). These 
%are adaptations from the models already presented in literature and 
models form the initial set $\delta$ in Algorithm~\ref{alg:boshnas}.

\begin{figure}[t]
    \centering
    \includegraphics[width=0.6\linewidth]{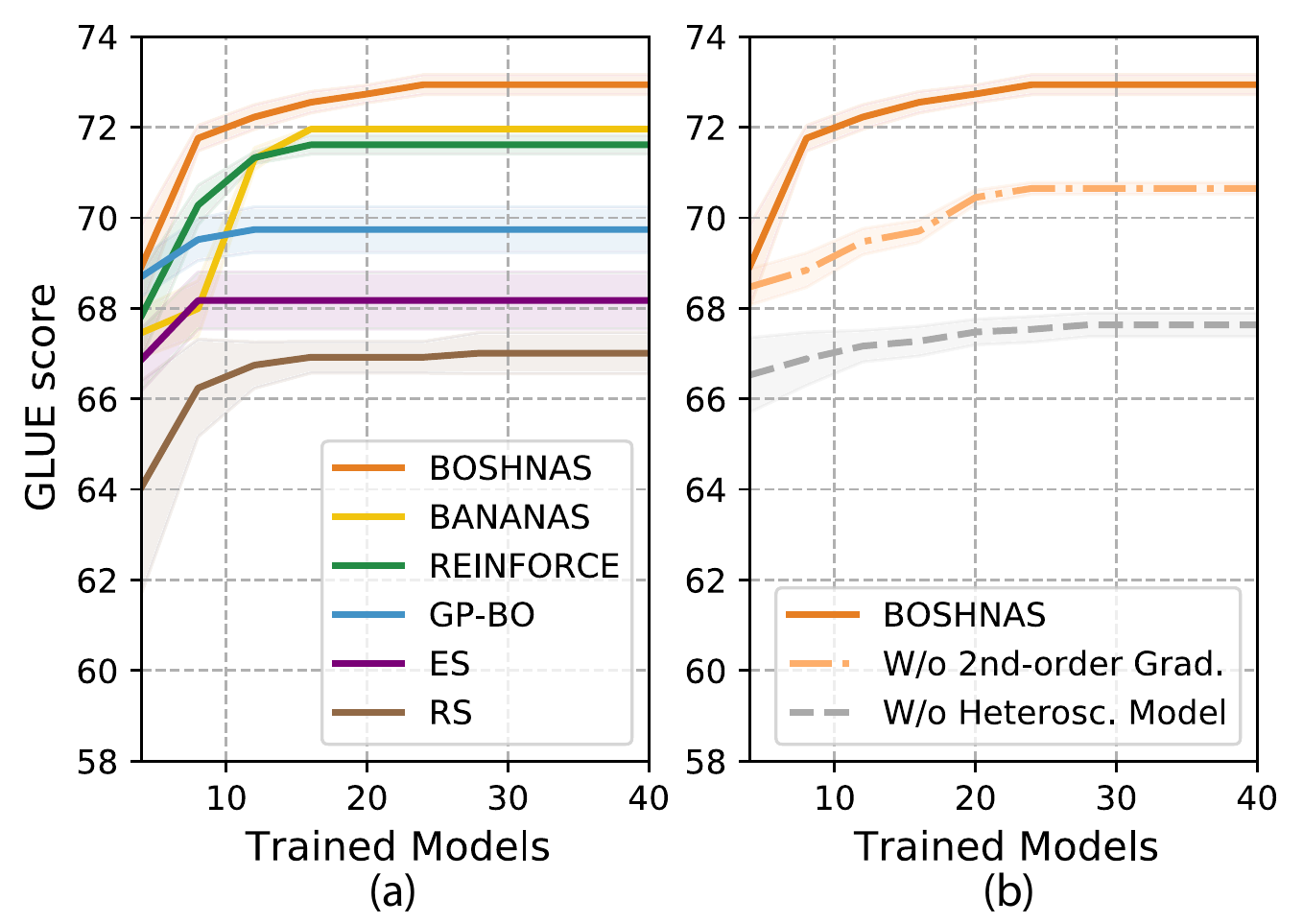}
    \caption{Performance results: (a) best GLUE score with trained models for NAS 
baselines and (b) ablation of BOSHNAS. Plotted with 90\% confidence intervals.}
    \label{fig:baselines_samples}
\end{figure}

\subsection{Ablation Study of BOSHNAS}
\label{sec:results_boshnas_ablation}

We compare BOSHNAS against other popular techniques 
from the CNN space, namely Random Search (RS), ES, REINFORCE, GP-BO, and a recent state-of-the-art, 
BANANAS. We present performance on the GLUE benchmark.

Figure~\ref{fig:ablation_baselines} presents the best GLUE scores reached by respective baseline 
NAS techniques along with BOSHNAS used with naive (\emph{i.e.}, feature-based one-hot) or 
\texttt{Transformer2vec} embeddings on a representative design space. We use the space in the first level 
of the hierarchy (\emph{i.e.}, with 9312 graphs, $s=2$) and run all these algorithms in an 
active-learning scenario (all targeted homogeneous models form a subset of this space) over 50 runs 
for each algorithm. The plot highlights that enhancing the richness of the design space enables the 
algorithms to search for more accurate models (6\% improvement averaged across all models). We also 
see that \texttt{Transformer2vec} embeddings help NAS algorithms reach better-performing architectures 
(9\% average improvement). Overall, BOSHNAS with the \texttt{Transformer2vec} embeddings performs the 
best in this representative design space, outperforming the state-of-the-art 
(\emph{i.e.}, BANANAS on naive embeddings) by \nasperf.

Figure~\ref{fig:baselines_samples}(a) shows the best GLUE score reached by each baseline NAS
algorithm along with the number of models it trained. Again, these runs are performed on the 
representative design space described above, using the \texttt{Transformer2vec} encodings. As can be seen from the figure, BOSHNAS reaches the best GLUE score. Ablation analysis justifies the need for heteroscedastic
modeling and second-order optimization (see Figure~\ref{fig:baselines_samples}(b)). The heteroscedastic model forces
the optimization of the training recipe when the framework approaches optimal
architectural design decisions. Second-order gradients, on the other hand, help the
search avoid local optima and saddle points, and also aid faster convergence.

\begin{table*}[]
\caption{Comparison between FlexiBERT and baselines. Results are evaluated on the development set of
the GLUE benchmark. We use Matthews correlation for CoLA, Spearman correlation for STS-B, and
accuracy for other tasks. MNLI is reported on the matched set. Ablation models for BOSHNAS
without second-order gradients (w/o S.) and without using the heteroscedastic model (w/o H.) are
also included. Best (second-best) performance values are in boldface (underlined). $^*$Performance for NAS-BERT$_{10}$ was not reported on the WNLI dataset and was reproduced using an equivalent model in our design space. $^\dagger$FlexiBERT-Mini model that only optimizes performance on the first eight tasks, for fair comparisons with NAS-BERT.}
\vskip 0.1in
\small
\centering
\resizebox{\linewidth}{!}{  
\begin{tabular}{@{}l|c|ccccccccc|c@{}}
\toprule
Model & Parameters & CoLA & MNLI & MRPC & QNLI & QQP & RTE & SST-2 & STS-B & WNLI & Avg. \\ \midrule
BERT-Mini~\cite{turc2019} & 16.6M & 0 & 74.8 & 71.8 & 84.1 & 66.4 & 57.9 & 85.9 & 73.3 & 62.3 & 64.0 \\
NAS-BERT$_{10}$~\cite{nas-bert} & 10M & \underline{27.8} & 76.0 & 81.5 & 86.3 & \underline{88.4} & 66.6 & \underline{88.6} & \textbf{84.8} & 53.7$^*$ & 72.6 \\ \midrule
FlexiBERT-Mini (ours, w/o S.) & \textbf{7.2M} & 16.7 & 72.3 & 72.9 & 81.7 & 76.9 & 64.1 & 80.9 & 77.0 & 65.3 & 67.5 \\
FlexiBERT-Mini (ours, w/o H.) & 20M & 12.3 & 74.4 & 72.3 & 76.4 & 76.3 & 59.5 & 81.2 & 75.4 & \underline{67.8} & 66.2 \\
FlexiBERT-Mini$^\dagger$ (ours) & 13.8M & \textbf{28.7} & \textbf{77.5} & \underline{82.3} & \underline{86.9} & 87.8 & \underline{67.6} & \textbf{89.7} & \underline{83.0} & 51.8 & \underline{72.7} \\
FlexiBERT-Mini (ours) & 16.1M & 23.8 & \underline{76.1} & \textbf{82.4} & \textbf{87.1} & \textbf{88.7} & \textbf{69.0} & 81.0 & 78.9 & \textbf{69.3} & \textbf{72.9} \\  \bottomrule
\end{tabular}}
\label{tab:baselines}
\end{table*}

Table~\ref{tab:baselines} shows the scores of the ablation models on the GLUE benchmarking tasks.
We refer to the best model obtained from BOSHNAS in the Tiny-to-Mini space as 
FlexiBERT-Mini. Once we
get the best architecture from the search process (using the same, albeit limited compute budget
for feasible search times), it is pre-trained and fine-tuned on a larger compute budget (details in 
Appendix~\ref{app:model_training}). As can be seen from the table, FlexiBERT-Mini outperforms the 
baseline, NAS-BERT~\shortcite{nas-bert}, by \perf on the GLUE benchmark. Since NAS-BERT finds the 
higher-performing architecture while only considering the first eight GLUE tasks (\emph{i.e.}, without 
the WNLI dataset), for fair comparisons, we find a neighboring model in the FlexiBERT design
space that only optimizes performance on the first eight tasks. We call this model
FlexiBERT-Mini$^\dagger$. We see that although FlexiBERT-Mini$^\dagger$ does not have the highest
GLUE score, it outperforms NAS-BERT$_{10}$ by significant margins on the first eights tasks.

\begin{figure}
    \centering
    \includegraphics[width=0.6\linewidth]{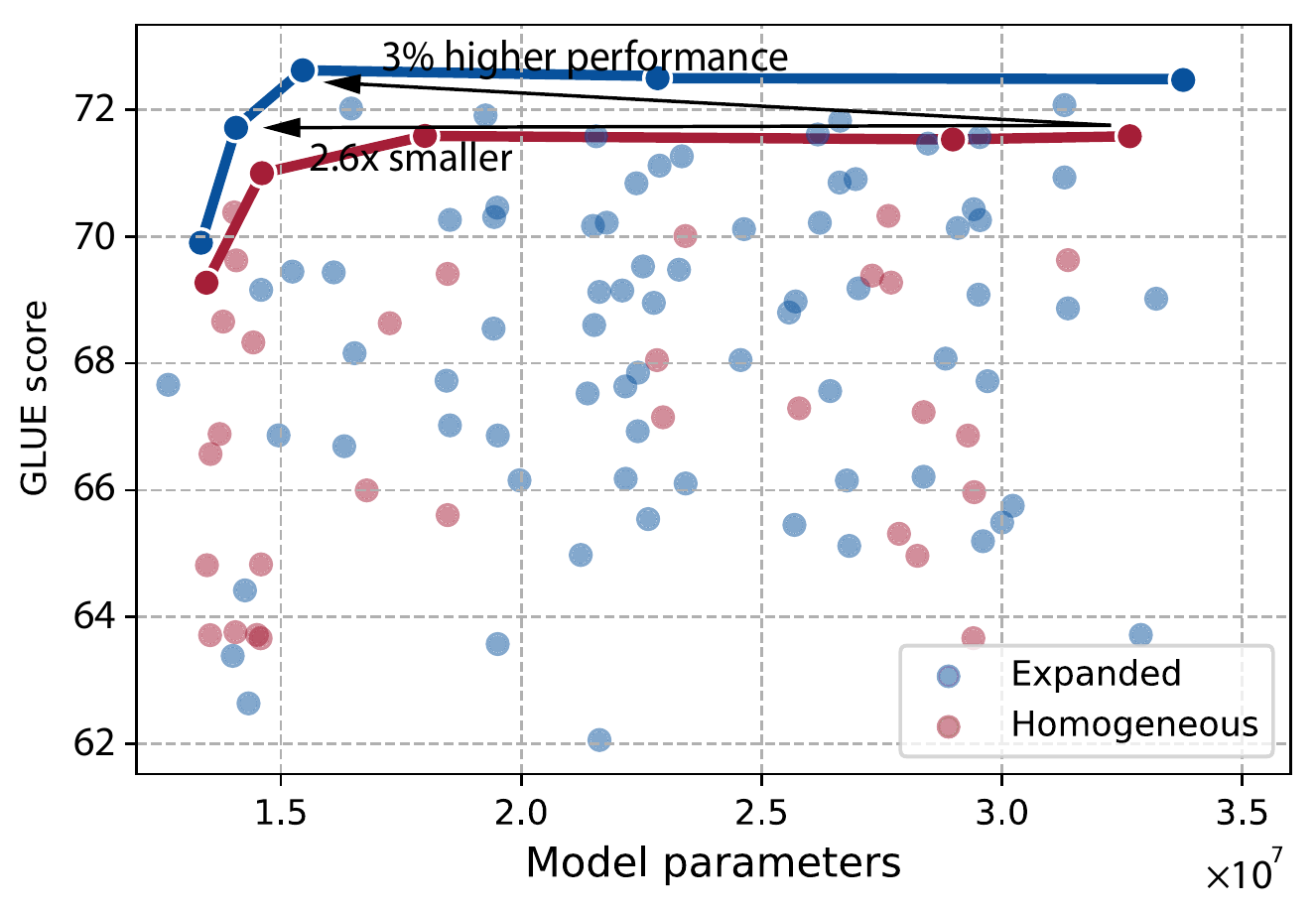}
    \caption{Performance frontiers of FlexiBERT on an expanded design space (under the constraints defined in Table~\ref{tab:hyp_ranges}) and for traditional homogeneous models.}
    \label{fig:frontier}
\end{figure}

Figure~\ref{fig:frontier} demonstrates 
that FlexiBERT pushes to improve the performance frontier relative to traditional homogeneous architectures. In other words, the best-performing models in the expanded (Tiny-to-Mini) space outperform 
traditional models for the same number of parameters. Here, the homogeneous models incorporate the 
same design decisions for all encoder layers, even with the expanded set of operations 
(\emph{i.e.}, including convolutional and LT-based attention operations). FlexiBERT-Mini has \parambertmini fewer parameters than BERT-Mini and achieves \perfbertmini higher GLUE score. FlexiBERT achieves \perfhomo higher performance than the best homogeneous model while the model with equivalent performance achieves \paramhomo smaller size.

\subsection{Best Architecture in the Design Space}
\label{sec:best_arch}

After running BOSHNAS for each level of the hierarchy, we get the respective best-performing 
models, whose model cards are presented in Appendix~\ref{app:best_models}. From these best-performing models, we 
can extract the following rules that lead to high-performing transformer architectures:
\begin{itemize}
    \item Models with DCT in the deeper layers are preferable for higher performance on the GLUE benchmark.
    \item Models with more attention heads, but smaller hidden dimension, are preferable in the deeper 
layers.
    \item Feed-forward networks with larger widths, but smaller depth, are preferable in the deeper 
layers.
\end{itemize}

\begin{table}[]
\caption{Comparison between FlexiBERT-Large (outside of the constraints defined in Table~\ref{tab:hyp_ranges}) and baselines on GLUE score. $^*$GLUE scores reported do not consider the WNLI dataset.}
\vskip 0.1in
\small
\centering
\resizebox{0.7\linewidth}{!}{  
\begin{tabular}{@{}lcc@{}}
\toprule
Model & Parameters & GLUE score \\ \midrule
RoBERTa~\shortcite{roberta} & 345M & 88.5  \\
FNet-Large~\shortcite{fnet} & 357M & 81.9$^*$ \\ \midrule
AutoTinyBERT~\shortcite{autotinybert} & 85M & 81.2$^*$ \\
DynaBERT~\shortcite{dynabert} & 345M & 81.6$^*$ \\
NAS-BERT$_{60}$~\shortcite{nas-bert} & 60M & 83.2$^*$ \\
AutoBERT-Zero Large~\shortcite{autobert_zero} & 318M & 84.5$^*$ \\ \midrule
FlexiBERT-Large (ours) & 319M & \textbf{89.1/90.2$^*$} \\ \bottomrule
\end{tabular}}
\label{tab:large_glue_scores}
\end{table}

Using these guidelines, we extrapolate the model card for FlexiBERT-Mini to get the design decisions for FlexiBERT-Large, which is an equivalent counterpart of BERT-Large~\shortcite{bert}. Appendix~\ref{app:best_models} presents the approach for extrapolation of hyperparameter choices from FlexiBERT-Mini to obtain FlexiBERT-Large. We train FlexiBERT-Large with the larger compute budget (see Appendix~\ref{app:model_training}) and show its GLUE score in Table~\ref{tab:large_glue_scores}. FlexiBERT-Large outperforms the baseline RoBERTa by 0.6\% on the entire GLUE benchmarking suite, and AutoBERT-Zero Large by \largeperf when only considering the first eight tasks.

\begin{figure}
    \centering
    \includegraphics[width=\linewidth]{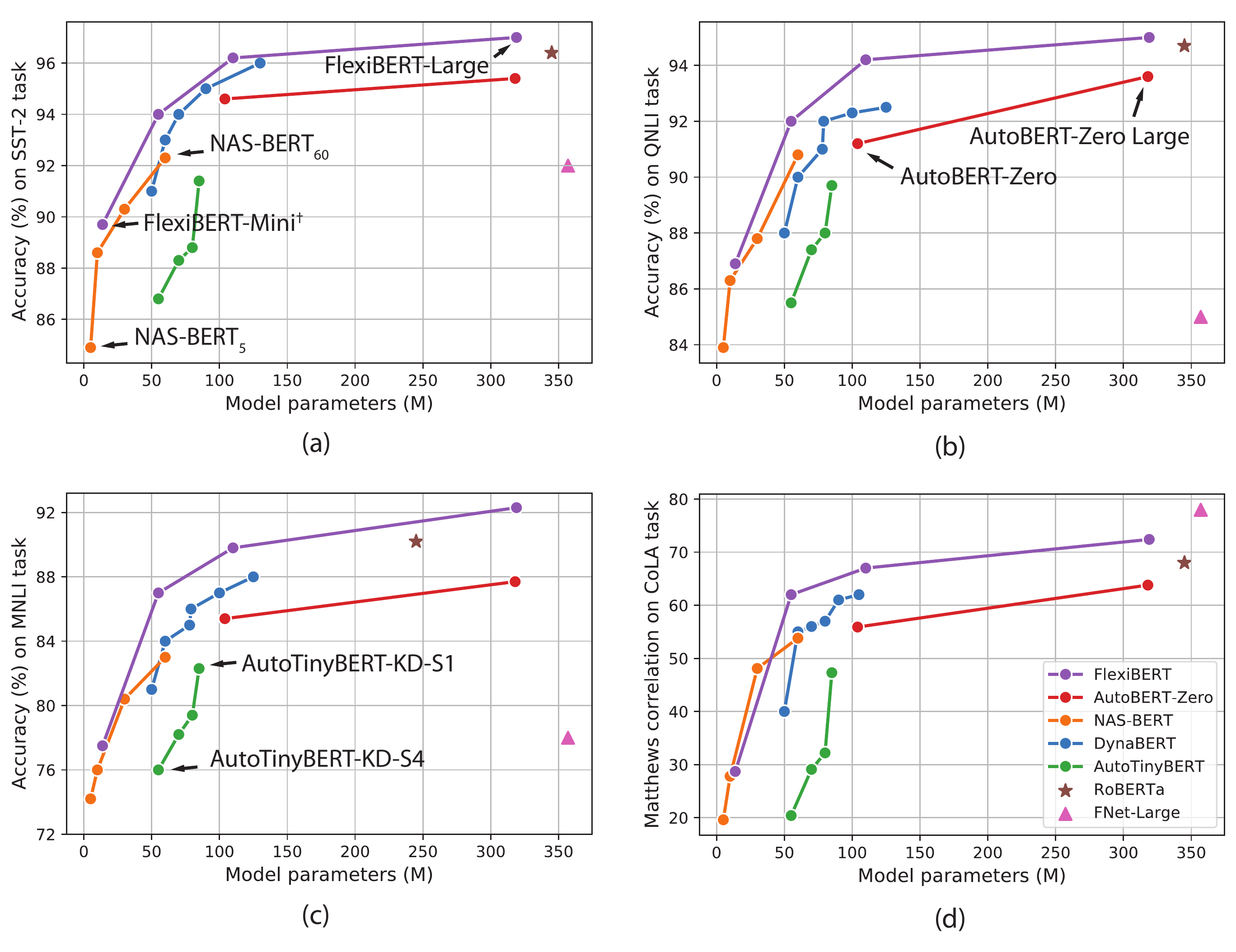}
    \caption{Performance of FlexiBERT and other baseline methods on various GLUE tasks: (a) SST-2, (b) QNLI, (c) MNLI (accuracy of MNLI-m is plotted), and (d) CoLA.}
    \label{fig:frontier_expanded}
\end{figure}

Just like FlexiBERT-Large is the BERT-Large counterpart of FlexiBERT-Mini, we 
similarly form the BERT-Small and BERT-Base equivalents~\shortcite{turc2019}. Figure~\ref{fig:frontier_expanded} presents 
the performance frontier of these FlexiBERT models with different baseline works. As can be seen, 
FlexiBERT consistently outperforms the baselines for different constraints on model size, thanks to 
its search in a vast, \emph{heterogeneous}, and \emph{flexible} design space of architectures.

\section{Conclusion}
\label{sec:conclusion}

In this work, we presented FlexiBERT, a suite of \emph{heterogeneous} and \emph{flexible} transformer models. 
We characterized the effects of this expanded design space and proposed a novel \texttt{Transformer2vec} 
embedding scheme to train a surrogate model that searches the design space for high-performance models. 
We described a novel NAS algorithm, BOSHNAS, and showed that it outperforms the state-of-the-art by 
\nasperf. The FlexiBERT-Mini model searched in this design space has a GLUE score that is \perfbertmini higher than BERT-Mini, while requiring \parambertmini lower parameters. It also beats the baseline, NAS-BERT$_{10}$ by 0.4\%. A FlexiBERT model with equivalent performance as the best homogeneous model achieves \paramhomo smaller size. FlexiBERT-Large outperforms the state-of-the-art models by at least \largeperf average accuracy on the first eight tasks in the GLUE benchmark.

\section*{Acknowledgments}

This work was supported by NSF Grant No. CNS-1907381 and CCF-2203399. The experiments reported in this 
paper were substantially performed on the computational resources managed and supported by 
Princeton Research Computing at Princeton University. We also thank Xiaorun Wu for initial discussions.

\newpage
\appendix

% \section*{\centering Appendix}

\section{Background}

Here, we discuss some supplementary background concepts.

\subsection{Self-Attention}
\label{app:self_attention}

Traditionally, transformers have relied on the SA operation. It is basically a
trainable associative memory. We depict the vanilla SA operation as SDP and introduce the WMA operation in our design space as well. For a source vector $\mathbf{s}$ and a hidden-state 
vector $\mathbf{h}$:
\begin{equation*}
    \text{SDP} := \frac{\mathbf{s}^\top \mathbf{h}}{\sqrt{d}}, \
    \text{WMA} := \mathbf{s}^\top \mathbf{W}_a \mathbf{h}
\end{equation*}
where $d$ is the dimension of the source vector and $\mathbf{W}_a$ is a trainable weight 
matrix in the attention layer. Naturally, a WMA layer is more expressive than an SDP layer.

The SA mechanism used in the context of transformers also involves the softmax
function and matrix multiplication. More concretely, in a multi-headed SA operation
(with $n$ heads), there are four matrices: $\mathbf{W}^q_i \in \mathbb{R}^{d_{inp} \times h/n}$, 
$\mathbf{W}^k_i \in \mathbb{R}^{d_{inp} \times h/n}$, $\mathbf{W}^v_i \in \mathbb{R}^{d_{inp} \times
h/n}$, and $\mathbf{W}^o_i \in \mathbb{R}^{h/n \times d_{out}}$, and it takes the hidden states of the previous layer as input $\mathbf{H} \in \mathbb{R}^{N_T \times d_{inp}}$, where $i$ refers to an attention head, $d_{inp}$ is the input dimension, $d_{out}$ is the output dimension, and $h$ is the hidden dimension. The output of 
the attention head ($\mathbf{H}_i \in \mathbb{R}^{N_T \times d_{out}}$) is then calculated as follows:
\begin{align*}
    \mathbf{Q}_i, \mathbf{K}_i, \mathbf{V}_i = \mathbf{H} \mathbf{W}^q_i, \mathbf{H} \mathbf{W}^k_i, \mathbf{H} \mathbf{W}^v_i \\
    \mathbf{H}_i = \text{softmax}\left( \frac{\mathbf{Q}_i \mathbf{K}_i^\intercal}{\sqrt{h}} \right) \mathbf{V}_i \mathbf{W}^o_i
\end{align*}
For traditional \emph{homogenous} transformer models, $d_{out}$ has to equal $d_{inp}$ (usually, $d_{inp} = d_{out} = h$) due to the residual connections. However, thanks to the \emph{relative} and \emph{trained} positional encodings and the added projection layer at the end of each encoder ($\mathbf{W}^p \in \mathbb{R}^{d_{out} \times d_p}$), we can relax this constraint. This leads to an expansion of the \emph{flexibility} of transformer models in the FlexiBERT design space.

\subsection{Improving BERT's performance}
\label{app:bert_improv}

BERT is one of the most widely used transformer architectures \shortcite{bert}.
Researchers have improved BERT's performance by revamping the pre-training technique. RoBERTa proposed a more 
robust pre-training approach to improve BERT's performance by considering \emph{dynamic} masking 
in the Masked Language Modeling (MLM) objective \shortcite{roberta}. Functional improvements 
have also been proposed for pre-training -- XLNet introduced Permuted Language Modeling (PLM) 
\shortcite{xlnet} and MPNet extended it by unifying MLM and PLM techniques \shortcite{mpnet}. Other 
approaches, including denoising autoencoders, have also been proposed \shortcite{bart}.

On the other hand, \shortciteA{schubert} consider optimizing the set of \emph{architectural}
design decisions for BERT -- number of encoder layers $l$, size of hidden embeddings $h$,
number of attention heads $a$, size of the hidden layer in the feed-forward network $f$,
etc. However, it is only concerned with pruning BERT and does not target optimization of
accuracy over different tasks. Further, it has a limited search space consisting of only
\emph{homogeneous} models.

\section{Experimental Details}

We present the details of the experiments performed next.

\subsection{Possible Compute Blocks}
\label{app:compute_blocks}

Based on the design space presented in Table~\ref{tab:hyp_ranges}, we consider all possible compute 
blocks, as presented next:
\begin{itemize}[nosep,itemindent=\parindent]
    \item For layer $j$, when the operation is SA, we have two or four heads 
with: $h$-128/SA-SDP, $h$-128/SA-WMA, $h$-256/SA-SDP, and $h$-256/SA-WMA. 
If the encoder layer has an LT operation, then we have two or four heads with: 
$h$-128/LT-DFT, $h$-128/LT-DCT, $h$-256/LT-DFT, and $h$-256/LT-DCT; the latter entry being 
the type of LT operation. For a \emph{convolutional} (DSC) operation, we have two or four 
heads with: $h$-128/DSC-5, $h$-128/DSC-9, $h$-256/DSC-5, and $h$-256/DSC-9; the latter entry referring to the 
kernel size.
    \item For layer $j$, the size of the hidden layer in the feed-forward network is either 
512 or 1024. Also, the feed-forward network may either have just one hidden layer or a stack 
of three layers. At higher levels of the hierarchy in the hierarchical search framework 
(details in Section~\ref{sec:comp_graph}), all the layers in the stack of hidden layers 
have the same dimension, until we relax this constraint in the last leg of the hierarchy. 
    \item Other blocks like: Add\&Norm, Input, and Output.
\end{itemize}

\subsection{Knowledge Transfer}
\label{app:know_transfer}

\begin{figure}
    \centering
    \includegraphics[width=0.6\linewidth]{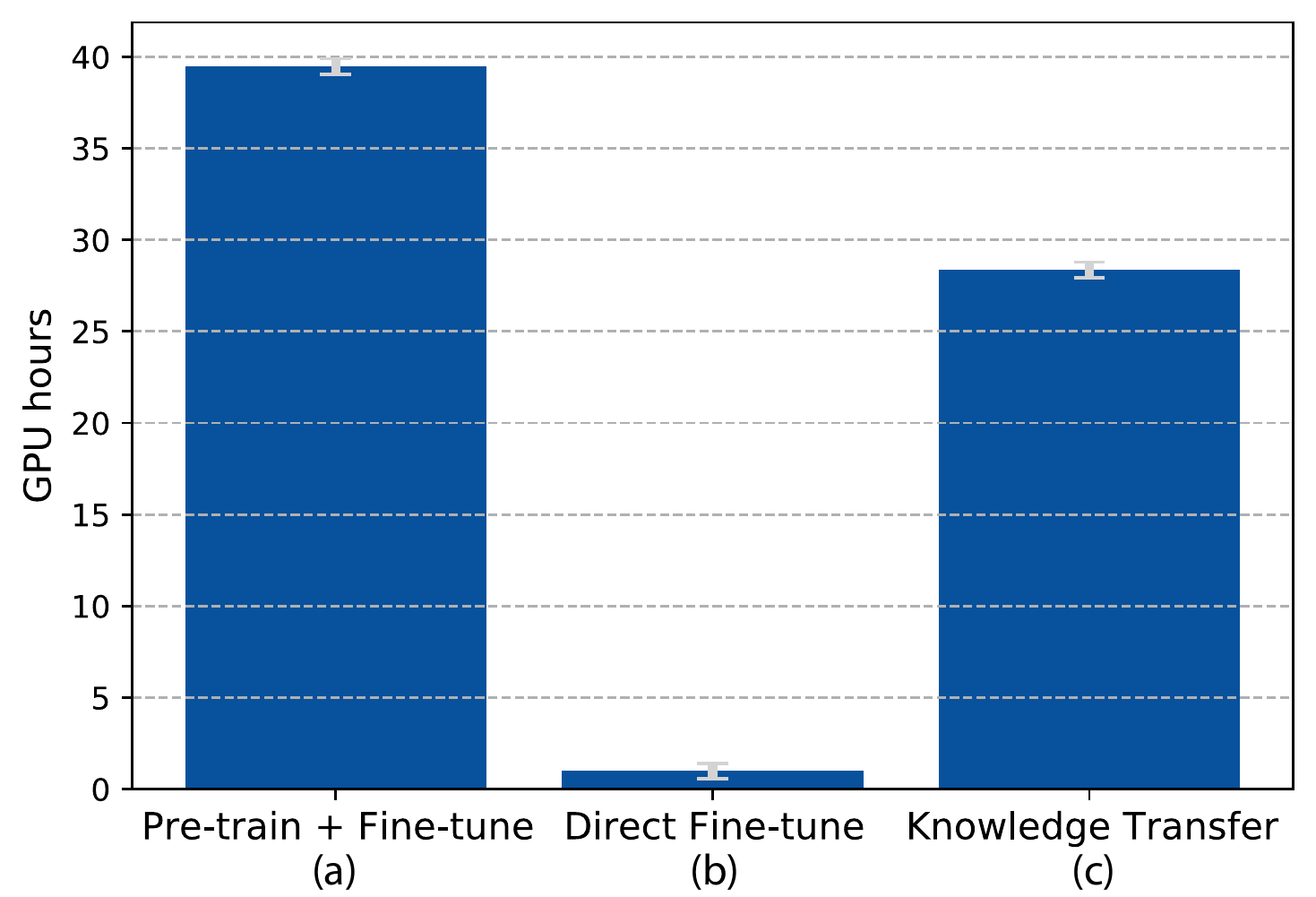}
    \caption{Bar plot showing average time for training a transformer model (in GPU-hours) with and without knowledge transfer. (a) Pre-train + Fine-tune: total training time. (b) Direct Fine-tune: training time for a pre-trained model. (c) Knowledge Transfer: training using weight transfer from a trained \emph{nearby} model, gives 38\% speedup. Plotted with 90\% confidence intervals.} %Computation time for an analogous approach without knowledge transfer was determined by statistical averages of times required for pre-trainining and fine-tuning of those models that were not directly fine-tuned.
    \label{fig:time_gains}
\end{figure}

Knowledge transfer has been used in recent works, but restricted to long short-term
memories and simple recurrent neural networks \shortcite{interspeech}. \shortciteA{hat_mit} train a 
super-transformer and share its weights with smaller models. However, this is not feasible 
for diverse \emph{heterogeneous} and \emph{flexible} architectures. We propose the use of knowledge transfer 
in transformers for the first time, to the best of our knowledge, by comparing weights with 
computational graphs of \emph{nearby} models. Furthermore, previous works only consider a static training recipe 
for all the models in the design space, an assumption we relax in our experiments.
We \emph{directly} fine-tune models for which \emph{nearby} models are already pre-trained. We test for 
this using the \emph{biased overlap} metric defined in Section~\ref{sec:transformer2vec}. 
Figure~\ref{fig:time_gains} presents the time gains from knowledge transfer. Since some percentage of 
models could directly be fine-tuned, thanks to their neighboring pre-trained models, we were able to 
speed up the overall training time by 38\%.

\begin{figure}
    \centering
    \includegraphics[width=0.65\linewidth]{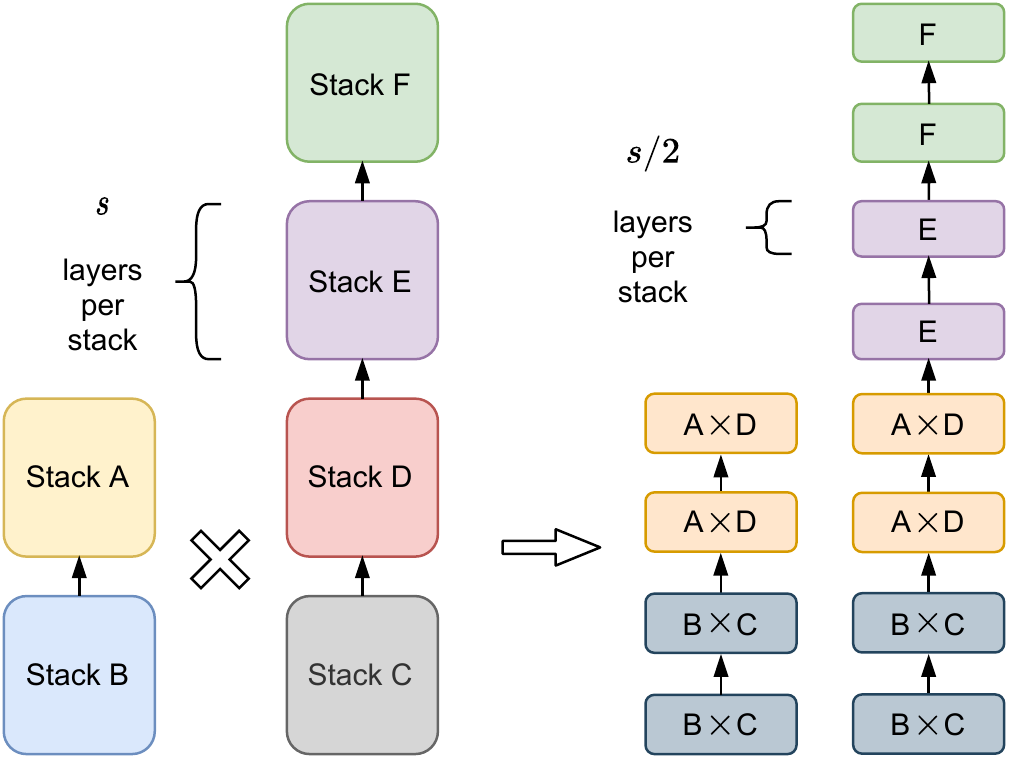}
    \caption{Crossover between two parent models yields a finer-grained design space. Each stack 
configuration in the children is derived from the product of the parent design choices at the 
same depth.}
    \label{fig:heirarchical}
\end{figure}

\subsection{Crossover between Transformer Models}
\label{app:crossover}

We obtain new transformer models of the subsequent level in the hierarchy by taking a crossover between the best models in the previous level (which had layers per stack $=s$) and their neighbors. The stack configuration of the children is chosen from all unique
hyperparameter values present in the parent models at the same depth. We present a simple
example of this scheme in Figure~\ref{fig:heirarchical}. The design space of permissible operation blocks for layers in the 
stack, $s$, is computed by the product of the respective design choices of the parents for that stack. These layers are then independently formed with the new constraint of $s/2$ layers having the same choice of hyperparameter values. Expanding the design space in such a 
fashion retains the original hyperparameters that give good performance while also exploring 
the internal representations learned by combinations of the hyperparameters at the same level.

\subsection{BOSHNAS Training Flow}
\label{app:boshnas_flow}

\begin{figure}[t]
    \centering
    \includegraphics[width=0.55\linewidth]{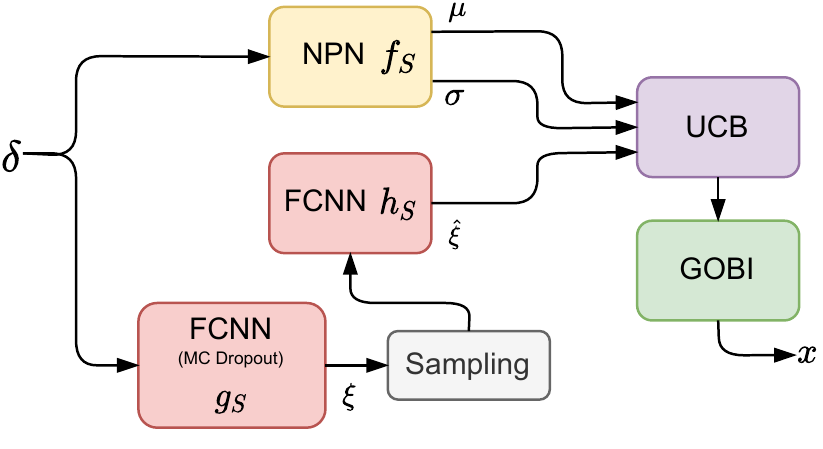}
    \caption{Overview of the BOSHNAS pipeline. Variables have been defined in Section~\ref{sec:boshnas}}
    \label{fig:boshnas_pipeline}
\end{figure}

Different surrogate models in the BOSHNAS pipeline ($f_S, g_S$, and $h_S$) have been presented in the order of flow in Figure~\ref{fig:boshnas_pipeline}. As explained in Section~\ref{sec:boshnas}, the NPN network ($f_S$) models the model performance and the \emph{aleatoric} uncertainty, and the student network ($h_S$) models the \emph{epistemic} uncertainty from the teacher network ($g_S$).

\subsection{Model Training}
\label{app:model_training}

We pre-train our models 
with a combination of publicly available text corpora, viz.~\texttt{BookCorpus} (BookC) 
\shortcite{bookcorpus}, \texttt{Wikipedia English} (Wiki), \texttt{OpenWebText} (OWT) \shortcite{OpenWebtext}, and 
\texttt{CC-News} (CCN) \shortcite{CC_news}. Most training hyperparameters are
borrowed from RoBERTa. We set the batch size to 256, learning rate warmed up over the
first $10,000$ steps to its peak value at $1 \times 10^{-5}$ that then decays linearly, weight decay to $0.01$, Adam scheduler's parameters $\beta_1 = 0.9$, 
$\beta_2 = 0.98$ (shown to improve stability; \shortciteR{roberta}), $\epsilon = 1 \times 10^{-6}$, and run 
pre-training for $1,000,000$ steps. 

Once the best models are found, we pre-train and fine-tune the selected models with a larger compute budget. For pre-training, we add the \texttt{C4} dataset \shortcite{c4} and train for $3,000,000$ steps before fine-tuning. We also fine-tune on each GLUE task for $10$ epochs instead of $5$ (further details are given below). This was done for the FlexiBERT-Mini and FlexiBERT-Large models. Table~\ref{tab:training_ablation} shows the improvement in performance of FlexiBERT-Mini that was trained using knowledge transfer (where the weights were transferred from a nearby trained model) after additional training. When compared to the model directly fine-tuned after knowledge transfer, we see only a marginal improvement when we pre-train from scratch. This reaffirms the advantage of knowledge transfer, that it reduces training time (see Appendix~\ref{app:know_transfer}) with negligible loss in performance. Training with a larger compute budget further improves performance on the GLUE benchmark, validating the importance of data size and diversity in pre-training~\shortcite{roberta}. Running a full-fledged BOSHNAS on the larger design space 
(\emph{i.e.}, with layers from 2 to 24, Tiny-to-Large) can be an easy extension of this work.

While running BOSHNAS, we fine-tune our models on the nine GLUE tasks over five epochs 
and a batch size of $64$ where early stopping is implemented. We also run automatic hyperparameter 
tuning for the fine-tuning process using the Tree-structured Parzen Estimator algorithm 
\shortcite{optuna_2019}. The learning rate is randomly selected logarithmically in the 
[$2\times 10^{-5},\;5 \times 10^{-4}$] range, and the batch size in $\{32, 64, 128\}$ uniformly. 
Table~\ref{tab:flexibert_training_recipe} shows the best hyperparameters for fine-tuning of each 
GLUE task selected using this auto-tuning technique. This hyperparameter optimization uses random 
initialization every time, which results in variation in performance each time the model is queried 
(see \emph{aleatoric} uncertainty explained in Section~\ref{sec:boshnas}).

\begin{table*}[]
\caption{Performance of FlexiBERT-Mini from BOSHNAS after knowledge transfer from a nearby trained model, and after pre-training from scratch along with a larger compute budget.}
\vskip 0.1in
\small
\centering
\resizebox{\linewidth}{!}{  
\begin{tabular}{@{}l|lcc|c@{}}
\toprule
Model & Pre-training data & Pre-training steps & Fine-tuning epochs & GLUE score \\ \midrule
FlexiBERT-Mini & & & \\
\hspace{3mm} w/ knowledge transfer & BookC, Wiki, OWT, CCN & 1,000,000 & 5 & 69.7 \\
\hspace{3mm} + pre-training from scratch & BookC, Wiki, OWT, CCN & 1,000,000 & 5 & 70.4 \\
\hspace{3mm} + larger compute budget & BookC, Wiki, OWT, CCN, C4 & 3,000,000 & 10 & 72.9 \\ \bottomrule
\end{tabular}}
\label{tab:training_ablation}
\end{table*}

\begin{table}
\small
\caption{Hyperparameters used for fine-tuning FlexiBERT-Mini on the GLUE tasks.}
\vskip 0.1in
\centering
\begin{tabular}{@{}lll@{}}
\toprule
Task & Learning rate & Batch size \\ \midrule
CoLA & $2.0 \times 10^{-4}$ & 64 \\
MNLI & $9.4 \times 10^{-5}$ & 64 \\
MRPC & $2.23 \times 10^{-5}$ & 32 \\
QNLI & $5.03 \times 10^{-5}$ & 128 \\
QQP & $3.7 \times 10^{-4}$ & 64 \\
RTE & $1.9 \times 10^{-4}$ & 128 \\
SST-2 & $1.2 \times 10^{-4}$ & 128 \\
STS-B & $7.0 \times 10^{-5}$ & 32 \\
WNLI & $4.0 \times 10^{-5}$ & 128 \\ \bottomrule
\end{tabular}
\label{tab:flexibert_training_recipe}
\end{table}

We have included baselines trained with the \emph{pre-training} + \emph{fine-tuning} procedure as
proposed by \shortciteA{turc2019} for like-for-like comparisons, and not the \emph{knowledge
distillation} counterparts~\shortcite{nas-bert}. Nevertheless, FlexiBERT is orthogonal to (and thus
can easily be combined with) knowledge distillation because FlexiBERT focuses on searching the
best architecture while knowledge distillation focuses on better training of a given architecture. 

All models were trained on NVIDIA A100 GPUs and 2.6 GHz AMD EPYC Rome processors. The entire process of running BOSHNAS for all levels of the hierarchy took around 300 GPU-days of training.

% We pre-train FlexiBERT models using the masked language model pre-training objective. However, recent works \shortcite{mpnet} 
% combine this with the Permutation Language Modeling (PLM) objective for autoregressive 
% prediction. FlexiBERT can therefore be trained using the PLM objective.  This would require 
% designing a two-stream attention mechanism for the convolution and 
% linear-transformation-based attention modules. We defer this to future work.

\subsection{Best-performing Models}
\label{app:best_models}

From different hierarchy levels ($s=2, 1$, and $1^*$), we get the respective best-performing models after running BOSHNAS as follows:
\begin{align*}
    s = 2 : \Big\{ l: 4, o: [\text{LT}, \text{LT}, \text{LT}, \text{LT}], h: [256, 256, 256, 256], n: [4, 4, 2, 2], \\ f: [[1024], [1024], [512, 512, 512], [512, 512, 512]], p: [\text{DCT}, \text{DCT}, \text{DCT}, \text{DCT}] \Big\} \\
    s = 1 : \Big\{ l: 4, o: [\text{SA}, \text{SA}, \text{LT}, \text{LT}], h: [256, 256, 128, 128], n: [2, 2, 4, 4], \\ f: [[512, 512, 512], [512, 512, 512], [1024], [1024]], p: [\text{SDP}, \text{SDP}, \text{DCT}, \text{DCT}] \Big\}
\end{align*}
where, in the last leg of the hierarchy, the stack length is $1$, but the feed-forward 
stacks are also heterogeneous (see Section~\ref{sec:comp_graph}). Both
$s=1$ and $s=1^*$ gave the same solution despite finer granularity in the
latter case. Thus, the second model card above is that of FlexiBERT-Mini.

The model cards of the FlexiBERT-Mini ablation models, as presented in Table~\ref{tab:baselines}, are given below:
\begin{align*}
    \text{(w/o S.)}: \Big\{ l: 2, o: [\text{SA}, \text{SA}], h: [128, 128], n: [4, 4], f: [[1024], [1024]], p: [\text{SDP}, \text{WMA}] \Big\} \\
    \text{(w/o H.)}: \Big\{ l: 4, o: [\text{LT}, \text{LT}, \text{SA}, \text{SA}], h: [256, 256, 128, 128], n: [4, 4, 4, 4], \\ f: [[1024, 1024, 1024], [1024, 1024, 1024], [512, 512, 512], [512, 512, 512]], \\ p: [\text{DCT}, \text{DCT}, \text{SDP}, \text{SDP}] \Big\}
\end{align*}

\begin{figure}[ht]
    \centering
    \includegraphics[width=0.7\linewidth]{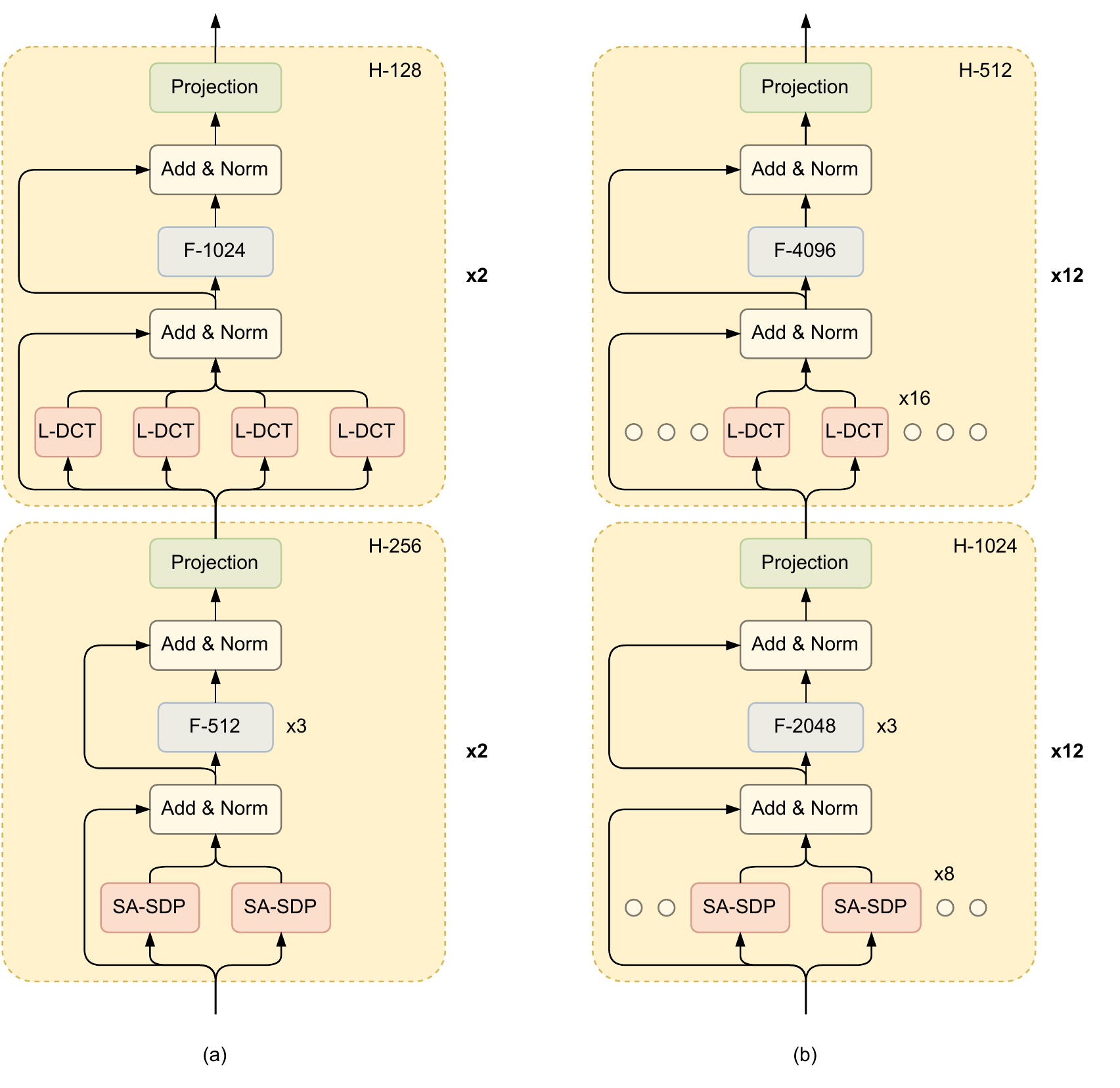}
    \caption{Obtained FlexiBERT models after running the BOSHNAS pipeline: (a) FlexiBERT-Mini, and its design choices extrapolated to obtain (b) FlexiBERT-Large.}
    \label{fig:flexibert_mini_large}
\end{figure}

Figure~\ref{fig:flexibert_mini_large} shows a working schematic of the design 
choices in the FlexiBERT-Mini and FlexiBERT-Large models. As explained in Section~\ref{sec:best_arch}, FlexiBERT-Large was formed by extrapolating the design choices in FlexiBERT-Mini to obtain a BERT-Large counterpart~\shortcite{bert}.

\clearpage
\vskip 0.2in
\bibliography{biblio}
\bibliographystyle{theapa}

\end{document}